\pdfoutput=1

\documentclass[11pt]{article}

\usepackage[]{acl}

\usepackage{times}
\usepackage{latexsym}
\usepackage[T1]{fontenc}
\usepackage[utf8]{inputenc}
\usepackage{microtype}
\usepackage{graphicx}
\usepackage{xcolor,soul}
\usepackage{booktabs}
\usepackage{enumitem}
\usepackage{etoolbox}
\usepackage{colortbl}
\usepackage{transparent}
\usepackage{amssymb}
\usepackage{url}

\newcommand*{\affaddr}[1]{#1}

\title{LLMs as Factual Reasoners: \\ Insights from Existing Benchmarks and Beyond}
\author{
  \quad \textbf{Philippe Laban}
  \quad \textbf{Wojciech Kry{\'s}ci{\'n}ski}
  \quad \textbf{Divyansh Agarwal}
  \quad \textbf{Alexander R. Fabbri} \\
  \quad \textbf{Caiming Xiong}
  \quad \textbf{Shafiq Joty}
  \quad \textbf{Chien-Sheng Wu} \\
  \affaddr{Salesforce AI} \\
  \{plaban, wojciech.kryscinski, dagarwal, afabbri, cxiong, sjoty, wu.jason\}@salesforce.com \\
}

\begin{document}

\newcommand\Tstrut{\rule{0pt}{2.8ex}}       % "top" strut
\newcommand\Bstrut{\rule[-1.4ex]{0pt}{0pt}} % "bottom" strut
\newcommand{\TBstrut}{\Tstrut\Bstrut} % top&bottom struts

% Edit-related commands
\definecolor{colordel}{HTML}{FECACA}
\definecolor{colorins}{HTML}{C7D9F7}

\definecolor{colorcorrect}{HTML}{87a0cf}
\definecolor{colorpartiallycorrect}{HTML}{ccdefa}
\definecolor{colornoexplanation}{HTML}{dedede}
\definecolor{colorunrelated}{HTML}{f2d6d5}
\definecolor{colorincorrect}{HTML}{d99493}

\DeclareRobustCommand{\hlred}[1]{{\sethlcolor{colordel}\hl{#1}}}
\DeclareRobustCommand{\hlblue}[1]{{\sethlcolor{colorins}\hl{#1}}}

\newcommand{\shafiq}[1]{\textcolor{cyan}{(shafiq: #1)}}
\newcommand{\jw}[1]{\textcolor{green}{\textbf{Jason:} #1}}
\newcommand{\wk}[1]{\textcolor{orange}{\textbf{Wojciech:} #1}}
% ["#02008b", "#6595ed", "#a0a0a0", "#b08080", "#b22322"]

\newcommand{\dataset}{\textsc{SummEdits}}

\maketitle
\begin{abstract}
With the recent appearance of LLMs in practical settings, having methods that can effectively detect factual inconsistencies is crucial to reduce the propagation of misinformation and improve trust in model outputs.
When testing on existing factual consistency benchmarks, we find that a few large language models (LLMs) perform competitively on classification benchmarks for factual inconsistency detection compared to traditional non-LLM methods. However, a closer analysis reveals that most LLMs fail on more complex formulations of the task and exposes issues with existing evaluation benchmarks, affecting the evaluation precision.
To address this, we propose a new protocol for inconsistency detection benchmark creation and implement it in a 10-domain benchmark called \dataset{}. This new benchmark is 20 times more cost-effective per sample than previous benchmarks and highly reproducible, as we estimate inter-annotator agreement at about 0.9.
Most LLMs struggle on \dataset{}, with performance close to random chance. The best-performing model, GPT-4, is still 8\% below estimated human performance, highlighting the gaps in LLMs' ability to reason about facts and detect inconsistencies when they occur.
% \jw{We argue about Inconsistency Detection not Inconsistency Generation. A model cannot identify inconsistency doesn’t explicitly mean that it will generate something inconsistent.}
\end{abstract}

\section{Introduction}

\begin{figure}
    \centering
    \includegraphics[width=0.48\textwidth]{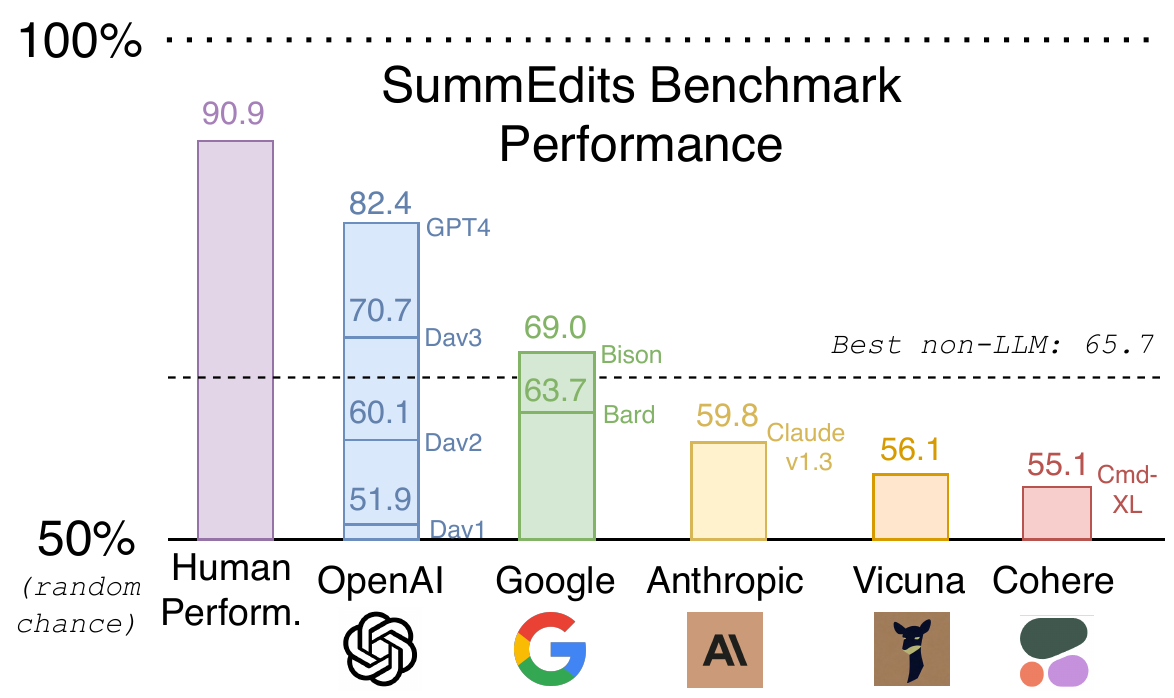}
    \caption{\dataset{} is a benchmark to evaluate the factual reasoning abilities of LLMs, measuring if models detect factual inconsistencies when they occur in summaries. Capable detection models can help build more reliable NLG systems.}
    \label{fig:benchmark_comparison}
\end{figure}

With recent progress in generation capabilities of LLMs, automatic summarization is making its appearance in practical information consumption situations such as summarizing work meetings \cite{arabzadeh2022preme}, health records \cite{Jain2022ASO}, or scientific documents \cite{cachola2020tldr}. To ensure the safe and effective implementation of these applications, it is essential to limit the reach of factually inconsistent summaries, a known issue with generated summaries \cite{kryscinski2019neural, maynez2020faithfulness}.

Prior work \cite{kryscinski2020evaluating,fabbri2021summeval,gao2022dialsummeval} has annotated corpora of model summaries with labels of factual consistency, finding that most abstractive summarization systems produce a non-negligible portion of inconsistent summaries. In turn, such corpora are used to instantiate tasks such as \textit{inconsistency detection} (ID) \cite{laban2022summac, tang2022understanding}, in which models are given \texttt{(document, summary)} pairs, and must identify whether the summary is consistent  with the document.

%Recent work has shown the promise of using LLMs for evaluation across NLP tasks \cite{liu2023gpteval,fu2023gptscore}, including for factual consistency \cite{Luo2023ChatGPTAA}.
Recent investigations of using LLMs for evaluation have shown promising results across different NLP tasks~\cite{liu2023gpteval,fu2023gptscore}, including factual consistency \cite{Luo2023ChatGPTAA}. In this work we continue this line of research and explore applying LLMs as factuality evaluators in the context of text summarization. We first establish baseline performance for a suite of LLMs on three existing consistency benchmarks. Accuracy results confirm that some LLMs perform competitively with state-of-the-art specialized methods such as QAFactEval \cite{fabbri2022qafacteval}. However, a manual analysis of free-text explanations that LLMs generate reveals two key limitations of the accuracy-only analysis. Ideally, a model correctly predicting the consistency label of a summary should be capable of generating an explanation for its binary prediction. Yet, we find that most LLMs generate explanations that do not accurately pinpoint factual inaccuracies, with only three models -- GPT4 \cite{OpenAI2023GPT4TR}, Claude V1.3 \cite{bai2022constitutional}, and Bard \cite{thoppilan2022lamda} -- generating correct explanations for more than 50\% of cases we annotated. Second, the manual analysis in the AggreFact consistency benchmark \cite{tang2022understanding} of conflict cases -- in which GPT4 predictions disagree with the dataset label -- reveals a significant number of mislabeled samples (7+\%) of factual inconsistencies undetected by annotators during dataset creation that the model explanation reveals. This lack of quality of benchmarks limits the precise evaluation of model performance at factual inconsistency detection.

To address this issue, we introduce a protocol designed to create challenging benchmarks while ensuring the reproducibility of the labels. The protocol involves manually verifying the consistency of a small set of seed summaries and subsequently generating numerous edited versions of these summaries. We discover that assessing the consistency of edited summaries is relatively straightforward and easy to scale for human annotators, thus guaranteeing low cost and high agreement among annotators, yet keeping the task challenging for models.

We create the \dataset{} benchmark by implementing the protocol in ten diverse textual domains, including the legal, dialogue, academic, financial, and sales domains. Figure~\ref{fig:benchmark_comparison} summarizes experimental results on the benchmark, which indicate that \dataset{} presents a challenge for both specialized models and current LLMs, with only four models — GPT3-Davinci003, ChatGPT, PaLM2-Bison, and GPT4 — outperforming the specialized model QAFactEval. Our estimate of human performance of 90\%+ is largely above all model performance, suggesting most current LLMs are not yet proficient at complex factual reasoning, and still cannot assess the factual validity of summaries with precision.

We believe \dataset{} can serve as a tool for evaluating LLMs' abilities to detect when factual inconsistencies (inevitably) occur and encourage LLM developers to report their performance on the benchmark. For practitioners requiring specific domain expertise, the protocol can be adapted to generate low-cost, in-domain benchmarks that can probe for model capabilities prior to production use. We release the code, protocol steps, and benchmark data publicly\footnote{\url{https://github.com/salesforce/factualNLG}}.

\section{Related Work}

\paragraph{Annotating Factuality of Summaries.} With advances in language models and the increase in fluency and abstractiveness of summarizers, prior work showed that one of the key challenges in summarization was enforcing factual consistency \cite{kryscinski2019neural}, particularly with models trained on datasets with unfaithful references \cite{maynez2020faithfulness}. Several efforts -- such as FactCC \cite{kryscinski2020evaluating}, SummEval \cite{fabbri2021summeval}, Polytope \cite{huang2020have}, FRANK \cite{pagnoni2021understanding}, and CLIFF \cite{cao2021cliff} -- annotated the generated summaries of tens of model, finding that most models produce a non-negligible portion of inconsistent summaries. Although most annotation effort has focused on the summarization of news, some prior work also looked at dialogue summarization \cite{gao2022dialsummeval}, or the medical domain \cite{tang2023evaluating}. In most work, scalable high-quality annotation is challenging, due to low inter-annotator agreement when relying on crowd-workers, with some work showing that 10+ annotators are required to achieve some level of consensus \cite{falke2019ranking}, and some work recommending solely relying on experts \cite{fabbri2021summeval}. At the heart of the issue, annotating the factual consistency of a summary is challenging: it requires careful reading of long documents and the detection and interpretation of nuanced facts. In this work, we propose a new protocol to annotate factual consistency resources and show that it lowers the cost and increases reproducibility by minimizing the amount of reasoning required for each annotation.

\paragraph{Detecting Factual Errors.} Some work has taken an automated approach to the detection of inconsistencies, with approaches falling into two main categories: question and entailment-based. In question-based approaches, questions are generated with the expectation that paired documents and summaries should provide consistent answers. QAFactEval \cite{fabbri2022qafacteval} unified prior work \cite{wang2020asking, scialom2021questeval} by systematically evaluating each element of the pipeline and proposing a best-performing combination. Entailment-based methods either rely on entailment on dependency parses, such as with the DAE method \cite{goyal2020evaluating}, or directly leverage natural-language entailment models, such as SummaC \cite{laban2022summac}. We include these three representative models in our experiments and find that although they require several orders of magnitudes fewer parameters than LLMs, they can reach similar performances on challenging benchmarks.

\section{LLM Aptitude In Controlled Setting}
\label{section:factcc_analysis}
In this section, we present the initial set of experiments that were conducted on the FactCC benchmark~\cite{kryscinski2020evaluating}. FactCC was created based on the XSum news summarization dataset~\cite{narayan2018xsum} and consists of news article-summary sentence pairs manually labeled based on their factuality. While simple in nature, the benchmark can serve as a test bed for exploring the basic understanding LLMs have of the task at hand. Furthermore, the data points come with manually annotated error types, allowing for experiments in fine-grained error detection.

In the following subsections, we define the experimental setup, i.e. prompts, models, and data, and present the experiment results along with a discussion.

\subsection{Prompt Selection}

% A variety of prompting techniques have been proposed as means of interacting with Large Language Models. However, there is no consensus within the research community on which of them consistently perform best across different LLM models.
As part of this initial study, we explore a wide range of prompts that have been shown to unlock some of the emergent abilities of LLMs. These prompts can be organized into four groups as follows:

\paragraph{Zero-Shot Prompts~\cite{radford2019language}} Evaluate the zero-shot transfer abilities of models. These prompts are limited to a short task description and the input data based on which the models generate their output. In our study, we included three different zero-shot prompts offering varying levels of detail in the task description. The best-performing prompt was selected by a majority vote across models and used as the base for the prompts described in the following paragraphs.

\paragraph{Few-Shot Prompts~\cite{brown2020language}} Enable the in-context learning abilities of LLMs. These prompts include a task description and one or more demonstrations of the task. The provided demonstrations condition the model for the actual input data that the model is expected to process. In this study we experiment with one-, two-, and three-shot prompts which build upon each other.

\paragraph{Chain-of-Thought Prompts~\cite{wei2022chain}} Explore LLM models' ability to generate step-by-step reasoning for answers and have been shown to improve performance on complex reasoning tasks. The models are given a task description and input data and are asked to generate a series of intermediate reasoning steps necessary to solve the task alongside the answer. We explore chain-of-thought prompts both in zero- and few-shot settings.

\paragraph{Generate-with-Evidence Prompts~\cite{lampinen2022language}} Explore the models' ability to present evidence for the generated answers and has also been shown to improve performance on reasoning-intense tasks. Similar to chain-of-thought prompts, the models are given a task description and input data and are asked to answer the task, and then generate evidence for the chosen answer. In this work we explore generate-with-evidence prompts in zero- and few-shot settings.

\paragraph{Persona-based Prompts~\cite{white2023prompt}} Extract certain points of view from LLMs or focus them on a set of abilities. Shown to work best with chat-tuned LLMs, models are assigned a role, or "persona", and next prompted to complete a given task. The assigned personas condition the models on the task at hand. In this work, models were assigned the persona of a ``journalist'' who is fact-checking a text before publication. Three prompts were tested, where the persona-based prompt was used in zero- and few-shot settings, and in combination with a chain-of-thought prompt.

% \paragraph{Fine-grained Evaluation Prompts} Previously described prompts were targeted at general factuality evaluation where the goal was to detect any (unspecified) type of factual inconsistency. To explore the model's understanding of factual correctness on a fine-grained level, we leverage the error-type labels assigned to FactCC examples. Fine-grained prompt query the models with evaluating factuality with respect to only one specific type of error, e.g. entity-related errors. Here, we explore individual prompts where each error type has an associated prompt. We also conduct experiments where individual prompts are combined with few-shot examples to aid the model in understanding the task.

All prompt templates described in this section and used in the study are presented in the associated code repository.

\subsection{Model Selection}
Similar to the prompt selection, we begin with evaluating a wide range of methods that can be applied to the task of factual consistency evaluation. Selected models span different architectures and training procedures, and can be categorized into the following groups:

\paragraph{Non-LLM} Models that were designed and trained specifically for the task of factual consistency evaluation in text summarization. Those models include two NLI-based approaches, DAE~\cite{goyal2020evaluating} and SummaC~\cite{laban2022summac}, and a QA-based method QAFactEval~\cite{fabbri2022qafacteval}. In this work, we treat the Non-LLM models as baselines and points of comparison with LLM-based factuality evaluators.

\paragraph{Foundation Models} Large-scale models that have been pre-trained on web-scale corpora, but have not been fine-tuned on task-specific or instruction-following datasets. Such models have shown emergent abilities, such as zero- and few-shot in-context learning. Models in this group include Meta's LLaMa-13b \cite{touvron2023llama}, and OpenAI's Ada001, Babbage001, Curie001, and DaVinci-001.

\paragraph{Instruction-tuned LLMs} Foundation models which were further tuned on instruction-following data either through supervised learning or RL-based methods. Such models show enhanced capabilities of following natural language instructions, including zero- and few-shot prompts as well as chain-of-thought approaches. Models in this group include Databrick's Dolly, Stanford's Alpaca \cite{taori2023alpaca}, Anthropic's Claude V1.3, Cohere's Command-XL, Google's PaLM2-bison, and OpenAI's DaVinci-002, and DaVinci-003 models.

\paragraph{Chat-based LLMs} Foundation models tuned on conversational and instruction-following datasets. The fine-tuning procedure aims to enhance the model capabilities of engaging in multi-turn dialog with end-users while being able to solve a wide range of complex tasks. This group includes Google's Bard, Mosaic's MPT-7b-chat \cite{MosaicML2023Introducing}, Vicuna-13b \cite{chiang2023vicuna}, and OpenAI's GPT3.5-turbo (ChatGPT), and GPT-4. 

For each model, model cards and method of access are provided in Appendix~\ref{appendix:model_access}, model architecture and training details are described in the associated literature.

\subsection{Experimental Setup}

Experiments described in the following subsections were conducted on the synthetic part of the FactCC dataset. We select a total of 150 samples to conduct experiments, by including 25 examples for each of the 5 error types in the dataset, i.e. date-, entity-, negation-, number-, and pronoun-related errors, and 25 factually correct samples. Considering that the original data was generated using heuristics, all examples were selected manually to ensure high-quality data.

\subsection{Inconsistency Detection}
\begin{table}[]
\centering
    \resizebox{0.45\textwidth}{!}{%
\begin{tabular}{llccccc}
\toprule
 & Model ($\downarrow$) & \multicolumn{5}{c}{Non-LLM Models} \\
\midrule
 \checkmark & DAE & \multicolumn{5}{c}{\cellcolor[rgb]{0.99, 0.99, 1.00} 67.2} \\
 \checkmark & SummaC & \multicolumn{5}{c}{ \cellcolor[rgb]{0.54, 0.56, 0.81} 96.8 } \\
 \checkmark & QAFactEval & \multicolumn{5}{c}{ \cellcolor[rgb]{0.57, 0.61, 0.85} 93.6 } \\
\midrule
 & Prompt Group $\rightarrow$ & ZS & FS & Pers & CoT & GwE \\
 \midrule
 & LLaMa-13B & \cellcolor[rgb]{0.85, 0.56, 0.56} 50.0 & \cellcolor[rgb]{0.86, 0.61, 0.61} 51.6 & \cellcolor[rgb]{0.87, 0.63, 0.63} 52.4 & - & - \\
 & Alpaca-13B & \cellcolor[rgb]{0.89, 0.69, 0.69} 54.8 & \cellcolor[rgb]{0.85, 0.56, 0.56} 48.4 & \cellcolor[rgb]{0.91, 0.75, 0.75} 57.2 & - & - \\
 & Dolly-v2-12B & \cellcolor[rgb]{0.85, 0.58, 0.58} 50.4 & \cellcolor[rgb]{0.86, 0.59, 0.59} 50.8 & \cellcolor[rgb]{0.86, 0.59, 0.59} 50.8 & - & - \\
 & MPT-7B-Chat & \cellcolor[rgb]{0.93, 0.79, 0.79} 58.7 & \cellcolor[rgb]{0.89, 0.67, 0.67} 54.0 & \cellcolor[rgb]{0.89, 0.68, 0.68} 54.4 & - & - \\
 \checkmark & Vicuna-13B & \cellcolor[rgb]{0.99, 0.97, 0.97} 65.5 & \cellcolor[rgb]{0.98, 0.98, 1.00} 68.0 & \cellcolor[rgb]{0.97, 0.91, 0.91} 63.2 & - & - \\
\checkmark & Cohere-CMD-XL & \cellcolor[rgb]{0.95, 0.86, 0.86} 61.3 & \cellcolor[rgb]{0.85, 0.56, 0.56} 50.0 & \cellcolor[rgb]{0.88, 0.65, 0.65} 53.3 & \cellcolor[rgb]{0.98, 0.95, 0.95} 64.7 & \cellcolor[rgb]{0.89, 0.69, 0.69} 54.8 \\
 \checkmark & Claude-v1.3 & \cellcolor[rgb]{0.82, 0.88, 0.98} 76.4 & \cellcolor[rgb]{0.69, 0.78, 0.96} 83.9 & \cellcolor[rgb]{0.90, 0.93, 0.99} 72.0 & \cellcolor[rgb]{0.76, 0.84, 0.97} 79.7 & \cellcolor[rgb]{0.81, 0.87, 0.98} 77.2 \\
 \checkmark & Bard & \cellcolor[rgb]{0.77, 0.84, 0.97} 79.3 & \cellcolor[rgb]{0.90, 0.93, 0.99} 72.3 & \cellcolor[rgb]{0.87, 0.91, 0.99} 73.7 & \cellcolor[rgb]{0.72, 0.81, 0.97} 82.0 & \cellcolor[rgb]{0.90, 0.93, 0.99} 71.9 \\
\checkmark & PaLM2-Bison & \cellcolor[rgb]{0.71, 0.80, 0.97} 82.3 & \cellcolor[rgb]{0.84, 0.89, 0.98} 75.5 & \cellcolor[rgb]{0.97, 0.92, 0.92} 63.7 & \cellcolor[rgb]{0.88, 0.92, 0.99} 73.1 & \cellcolor[rgb]{0.92, 0.94, 0.99} 71.3 \\
 & Ada001 & \cellcolor[rgb]{0.85, 0.56, 0.56} 46.4 & \cellcolor[rgb]{0.85, 0.56, 0.56} 47.7 & \cellcolor[rgb]{0.85, 0.56, 0.56} 49.6 & \cellcolor[rgb]{0.87, 0.62, 0.62} 52.0 & \cellcolor[rgb]{0.85, 0.56, 0.56} 50.0 \\
 & Bab001 & \cellcolor[rgb]{0.87, 0.61, 0.61} 51.9 & \cellcolor[rgb]{0.91, 0.75, 0.75} 57.1 & \cellcolor[rgb]{0.85, 0.56, 0.56} 49.5 & \cellcolor[rgb]{0.85, 0.56, 0.56} 49.1 & \cellcolor[rgb]{0.88, 0.65, 0.65} 53.1 \\
 & Cur001 & \cellcolor[rgb]{0.88, 0.66, 0.66} 53.5 & \cellcolor[rgb]{0.88, 0.65, 0.65} 53.3 & \cellcolor[rgb]{0.86, 0.59, 0.59} 51.1 & \cellcolor[rgb]{0.92, 0.76, 0.76} 57.5 & \cellcolor[rgb]{0.91, 0.73, 0.73} 56.3 \\
 \checkmark & Dav001 & \cellcolor[rgb]{0.95, 0.86, 0.86} 61.2 & \cellcolor[rgb]{0.91, 0.74, 0.74} 56.8 & \cellcolor[rgb]{0.88, 0.64, 0.64} 52.9 & \cellcolor[rgb]{0.95, 0.87, 0.87} 61.6 & \cellcolor[rgb]{0.92, 0.78, 0.78} 58.1 \\
 \checkmark & Dav002 & \cellcolor[rgb]{0.86, 0.90, 0.98} 74.5 & \cellcolor[rgb]{0.73, 0.82, 0.97} 81.3 & \cellcolor[rgb]{0.92, 0.77, 0.77} 57.9 & \cellcolor[rgb]{0.78, 0.85, 0.98} 78.5 & \cellcolor[rgb]{0.88, 0.92, 0.99} 73.2 \\
 \checkmark & Dav003 & \cellcolor[rgb]{0.71, 0.80, 0.97} 82.3 & \cellcolor[rgb]{0.79, 0.85, 0.98} 78.4 & \cellcolor[rgb]{0.96, 0.89, 0.89} 62.4 & \cellcolor[rgb]{0.67, 0.75, 0.94} 85.5 & \cellcolor[rgb]{0.81, 0.87, 0.98} 76.8 \\
 \checkmark & GPT3.5-turbo & \cellcolor[rgb]{0.68, 0.77, 0.95} 84.3 & \cellcolor[rgb]{0.70, 0.80, 0.97} 82.9 & \cellcolor[rgb]{0.85, 0.89, 0.98} 75.1 & \cellcolor[rgb]{0.69, 0.78, 0.96} 84.0 & \cellcolor[rgb]{0.66, 0.74, 0.93} 86.3 \\
 \checkmark & GPT4 & \cellcolor[rgb]{0.60, 0.65, 0.87} 91.3 & \cellcolor[rgb]{0.62, 0.67, 0.89} 90.1 & \cellcolor[rgb]{1.00, 0.99, 0.99} 66.3 & \cellcolor[rgb]{0.67, 0.75, 0.94} 85.7 & \cellcolor[rgb]{0.79, 0.86, 0.98} 78.0 \\
\bottomrule
\end{tabular}

    }
    \caption{Balanced accuracy on the synthetic FactCC benchmark per prompt group (averaged across prompts in each group). Specialized non-LLMs, (top) Foundation Models, Instruction-tuned LLMs, and Chat-based LLMs (bottom). For LLMs, performance is evaluated with Zero-shot (\texttt{ZS}), Few-Shot (\texttt{FS}), Persona (\texttt{Pers}), Chain-of-Thought (\texttt{CoT}), and Generate-with-Evidence (\texttt{GwE}) prompts when sequence length allows.}
    \label{table:factcc-bacc}
\end{table}

We first evaluate the models' ability to detect factual inconsistencies in a binary classification setting with \texttt{Yes}-\texttt{No} labels.
Non-LLM models return a continuous score attributed to a label class using a tuned threshold, while LLM-based approaches generate free-form text where the final output is retrieved using regular expressions. Due to input length restrictions, certain models could not be tested on more complex (and longer) prompts. Table~\ref{table:factcc-bacc} presents the balanced accuracy scores averaged across three prompts within each prompt category. The results provide the following insights:
\begin{table}[]
\centering
    \resizebox{0.47\textwidth}{!}{%
\begin{tabular}{lrrrrrr}
\cmidrule{3-7}
& & \multicolumn{5}{c}{Error Type} \\
\midrule
Model ($\downarrow$) & POS & DS & ES & NSent & NS & PS \\
\midrule
 DAE & \cellcolor[rgb]{0.55, 0.57, 0.82} 96.0 & \cellcolor[rgb]{0.85, 0.56, 0.56} 12.0 & \cellcolor[rgb]{0.85, 0.56, 0.56} 44.0 & \cellcolor[rgb]{0.85, 0.56, 0.56} 28.0 & \cellcolor[rgb]{0.87, 0.62, 0.62} 52.0 & \cellcolor[rgb]{0.85, 0.56, 0.56} 44.0 \\
 SummaC & \cellcolor[rgb]{0.55, 0.57, 0.82} 96.0 & \cellcolor[rgb]{0.50, 0.50, 0.78} 100.0 & \cellcolor[rgb]{0.50, 0.50, 0.78} 100.0 & \cellcolor[rgb]{0.50, 0.50, 0.78} 100.0 & \cellcolor[rgb]{0.50, 0.50, 0.78} 100.0 & \cellcolor[rgb]{0.76, 0.83, 0.97} 80.0 \\
 QAFactEval & \cellcolor[rgb]{0.55, 0.57, 0.82} 96.0 & \cellcolor[rgb]{0.69, 0.78, 0.96} 84.0 & \cellcolor[rgb]{0.59, 0.64, 0.87} 92.0 & \cellcolor[rgb]{0.55, 0.57, 0.82} 96.0 & \cellcolor[rgb]{0.55, 0.57, 0.82} 96.0 & \cellcolor[rgb]{0.69, 0.78, 0.96} 84.0 \\
\midrule
 LLaMa-13B & \cellcolor[rgb]{0.63, 0.69, 0.90} 88.8 & \cellcolor[rgb]{0.85, 0.56, 0.56} 10.4 & \cellcolor[rgb]{0.85, 0.56, 0.56} 13.6 & \cellcolor[rgb]{0.85, 0.56, 0.56} 14.4 & \cellcolor[rgb]{0.85, 0.56, 0.56} 12.8 & \cellcolor[rgb]{0.85, 0.56, 0.56} 12.8 \\
 Alpaca-13B & \cellcolor[rgb]{0.76, 0.83, 0.97} 80.0 & \cellcolor[rgb]{0.85, 0.56, 0.56} 30.4 & \cellcolor[rgb]{0.85, 0.56, 0.56} 20.0 & \cellcolor[rgb]{0.85, 0.56, 0.56} 36.0 & \cellcolor[rgb]{0.85, 0.56, 0.56} 25.6 & \cellcolor[rgb]{0.85, 0.56, 0.56} 28.0 \\
 Dolly-v2-12B & \cellcolor[rgb]{0.57, 0.61, 0.85} 93.6 & \cellcolor[rgb]{0.85, 0.56, 0.56} 3.2 & \cellcolor[rgb]{0.85, 0.56, 0.56} 11.2 & \cellcolor[rgb]{0.85, 0.56, 0.56} 10.4 & \cellcolor[rgb]{0.85, 0.56, 0.56} 7.2 & \cellcolor[rgb]{0.85, 0.56, 0.56} 5.6 \\
 MPT-7B & \cellcolor[rgb]{0.90, 0.93, 0.99} 72.0 & \cellcolor[rgb]{0.85, 0.56, 0.56} 36.0 & \cellcolor[rgb]{0.85, 0.56, 0.56} 41.6 & \cellcolor[rgb]{0.88, 0.64, 0.64} 52.8 & \cellcolor[rgb]{0.85, 0.56, 0.56} 38.4 & \cellcolor[rgb]{0.85, 0.56, 0.56} 40.0 \\
 Vicuna-13B & \cellcolor[rgb]{0.96, 0.97, 1.00} 68.8 & \cellcolor[rgb]{0.93, 0.81, 0.81} 59.2 & \cellcolor[rgb]{0.97, 0.91, 0.91} 63.2 & \cellcolor[rgb]{0.86, 0.90, 0.98} 74.4 & \cellcolor[rgb]{0.99, 0.97, 0.97} 65.6 & \cellcolor[rgb]{0.85, 0.56, 0.56} 48.8 \\
Cohere-CMD-XL & \cellcolor[rgb]{0.67, 0.76, 0.94} 85.1 & \cellcolor[rgb]{0.85, 0.56, 0.56} 32.0 & \cellcolor[rgb]{0.85, 0.56, 0.56} 31.5 & \cellcolor[rgb]{0.85, 0.56, 0.56} 36.3 & \cellcolor[rgb]{0.85, 0.56, 0.56} 26.1 & \cellcolor[rgb]{0.85, 0.56, 0.56} 17.1 \\
 Claude v1.3 & \cellcolor[rgb]{0.91, 0.94, 0.99} 71.7 & \cellcolor[rgb]{0.71, 0.80, 0.97} 82.4 & \cellcolor[rgb]{0.75, 0.83, 0.97} 80.3 & \cellcolor[rgb]{0.63, 0.69, 0.90} 89.1 & \cellcolor[rgb]{0.62, 0.68, 0.89} 89.9 & \cellcolor[rgb]{0.79, 0.86, 0.98} 78.1 \\
 Bard & \cellcolor[rgb]{0.76, 0.83, 0.97} 80.0 & \cellcolor[rgb]{0.96, 0.97, 1.00} 68.8 & \cellcolor[rgb]{0.95, 0.97, 0.99} 69.3 & \cellcolor[rgb]{0.81, 0.87, 0.98} 77.3 & \cellcolor[rgb]{0.69, 0.78, 0.96} 83.7 & \cellcolor[rgb]{0.93, 0.81, 0.81} 59.2 \\
 Palm2 & \cellcolor[rgb]{0.54, 0.56, 0.81} 96.5 & \cellcolor[rgb]{0.85, 0.56, 0.56} 47.7 & \cellcolor[rgb]{0.85, 0.56, 0.56} 45.3 & \cellcolor[rgb]{0.99, 0.96, 0.96} 65.1 & \cellcolor[rgb]{0.87, 0.62, 0.62} 52.0 & \cellcolor[rgb]{0.85, 0.56, 0.56} 38.9 \\
 Ada001 & \cellcolor[rgb]{0.93, 0.79, 0.79} 58.7 & \cellcolor[rgb]{0.85, 0.56, 0.56} 36.5 & \cellcolor[rgb]{0.85, 0.56, 0.56} 40.3 & \cellcolor[rgb]{0.85, 0.56, 0.56} 45.9 & \cellcolor[rgb]{0.85, 0.56, 0.56} 39.2 & \cellcolor[rgb]{0.85, 0.56, 0.56} 36.3 \\
 Bab001 & \cellcolor[rgb]{0.94, 0.96, 0.99} 70.1 & \cellcolor[rgb]{0.85, 0.56, 0.56} 33.9 & \cellcolor[rgb]{0.85, 0.56, 0.56} 29.6 & \cellcolor[rgb]{0.85, 0.56, 0.56} 41.6 & \cellcolor[rgb]{0.85, 0.56, 0.56} 30.7 & \cellcolor[rgb]{0.85, 0.56, 0.56} 34.7 \\
 Cur001 & \cellcolor[rgb]{0.64, 0.71, 0.91} 88.0 & \cellcolor[rgb]{0.85, 0.56, 0.56} 12.0 & \cellcolor[rgb]{0.85, 0.56, 0.56} 17.3 & \cellcolor[rgb]{0.85, 0.56, 0.56} 45.1 & \cellcolor[rgb]{0.85, 0.56, 0.56} 16.5 & \cellcolor[rgb]{0.85, 0.56, 0.56} 12.3 \\
 Dav001 & \cellcolor[rgb]{0.64, 0.71, 0.91} 88.0 & \cellcolor[rgb]{0.85, 0.56, 0.56} 21.9 & \cellcolor[rgb]{0.85, 0.56, 0.56} 28.5 & \cellcolor[rgb]{0.85, 0.58, 0.58} 50.4 & \cellcolor[rgb]{0.85, 0.56, 0.56} 27.5 & \cellcolor[rgb]{0.85, 0.56, 0.56} 13.1 \\
 Dav002 & \cellcolor[rgb]{0.75, 0.83, 0.97} 80.3 & \cellcolor[rgb]{1.00, 0.99, 0.99} 66.4 & \cellcolor[rgb]{0.95, 0.86, 0.86} 61.3 & \cellcolor[rgb]{0.85, 0.90, 0.98} 74.9 & \cellcolor[rgb]{0.91, 0.94, 0.99} 71.5 & \cellcolor[rgb]{0.90, 0.71, 0.71} 55.5 \\
 Dav003 & \cellcolor[rgb]{0.58, 0.62, 0.85} 93.1 & \cellcolor[rgb]{0.99, 0.99, 0.99} 66.1 & \cellcolor[rgb]{0.93, 0.78, 0.78} 58.4 & \cellcolor[rgb]{0.92, 0.94, 0.99} 71.2 & \cellcolor[rgb]{0.94, 0.96, 0.99} 69.9 & \cellcolor[rgb]{0.85, 0.56, 0.56} 39.7 \\
 GPT3.5-turbo & \cellcolor[rgb]{0.65, 0.72, 0.92} 87.2 & \cellcolor[rgb]{0.71, 0.80, 0.97} 82.4 & \cellcolor[rgb]{0.97, 0.92, 0.92} 63.5 & \cellcolor[rgb]{0.65, 0.72, 0.92} 87.5 & \cellcolor[rgb]{0.63, 0.69, 0.90} 89.1 & \cellcolor[rgb]{1.00, 1.00, 1.00} 66.7 \\
 GPT4 & \cellcolor[rgb]{0.66, 0.74, 0.93} 86.1 & \cellcolor[rgb]{0.85, 0.90, 0.98} 74.9 & \cellcolor[rgb]{0.81, 0.87, 0.98} 77.3 & \cellcolor[rgb]{0.73, 0.81, 0.97} 81.6 & \cellcolor[rgb]{0.68, 0.77, 0.95} 84.3 & \cellcolor[rgb]{0.86, 0.91, 0.98} 74.1 \\
\midrule
LLM Avg. & 81.64 & 44.95 & 44.25 & 56.10 & 48.81 & 38.87 \\
\bottomrule
\end{tabular}

    }
    \caption{Accuracy on the synthetic FactCC benchmark per error type (averaged across all prompts). Specialized non-LLMs (top) Foundation Models, Instruction-tuned LLMs, and Chat-based LLMs (bottom). Performance is assessed individually for positive examples (\texttt{POS}) and each of the error types: Date Swap (\texttt{DS}), Entity Swap (\texttt{ES}), Negated Sentences (\texttt{NSent}), Number Swap (\texttt{NS}), Pronoun Swap (\texttt{PS}).} 
    \label{table:factcc-bacc-errortype}
\end{table}

% 1st insight
Two non-LLM models achieve near-perfect accuracy and substantially outperform LLM-based evaluators. We speculate that this might be caused by non-LLM models being over-optimized to the idiosyncrasies of this simple error detection task and might not hold for more involved detection examples. We investigate this question further in later sections of the paper.

% 2nd insight
Regarding the prompt design, we notice that for most models (8/12), providing a few examples of the task (zero- $\rightarrow$ few-shot) improves the performance by an average of 2.7 percentage points. However, for two models, GPT4 and PaLM2, the performance in the same setting dropped substantially (-6.2 pp). Considering those the two models achieve strong performance across all prompt types, we conclude that few-shot examples can help models but are not necessary for top-performing models.

% 3rd insight
In the majority of cases (8/12) Generate-with-Evidence prompts outperform Chain-of-Thought prompts corroborating prior findings of~\citet{ye2022unreliability} that models perform better when generating an answer followed by the evidence, rather than generating reasoning followed by an answer as in CoT. An in-depth evaluation of the factual reasoning capabilities of models is presented in the following section. 

% 4th insight
Persona-based prompts improve the performance of GPT3.5-Turbo; however, they lower the performance of all other models, including the remaining chat-based LLMs. This finding suggests that conditioning the generation on a specific persona might be a feature exclusive to ChatGPT, possibly linked to the data used for tuning, rather than a broader trait of chat-tuned LLMs.

% 5th insight
We study model performance for each error type, averaging the accuracy score across all prompts for LLM models. Results are presented in Table~\ref{table:factcc-bacc-errortype}. We notice that the majority of LLM-based evaluators achieve satisfactory accuracy (> 80\%) in classifying positive (factually correct) examples. The results also highlight that with a few exceptions (Claude, ChatGPT, and GPT4) LLMs struggle to consistently detect factual inconsistencies, in many cases achieving scores below random chance. Pronoun swap detection seems to be the most complicated of error types averaging 38.87\% across all LLM-based evaluators and also scoring lowest for specialized, non-LLM models.

\subsection{Factual Reasoning}
\label{section:factcc_reasoning}

\begin{table}[]
    \centering
    \renewcommand{\arraystretch}{1} 
    \resizebox{0.44\textwidth}{!}{%
    \begin{tabular}{p{8cm}}
    \toprule
    \cellcolor[rgb]{0.97, 0.97, 0.97}
    \hspace{6.7em} \textbf{Input Document} \\ 
    \midrule
    Prue Leith, who takes over from \textbf{Mary Berry} on the show, insists ``baking is the best way to get people into cooking''. The latest series returns this autumn on Channel 4 after it outbid the BBC. The previous hosts announced they ``were not going with the dough'' when the show switched channels. The pair had presented all seven series of the show alongside judges Berry and Hollywood. Similarly, Berry also announced \textbf{she} would not remain with the show on Channel 4, citing "loyalty" to the BBC. [...] \\
    \midrule
    \cellcolor[rgb]{0.97, 0.97, 0.97}
    \hspace{6.0em} \textbf{Inconsistent Summary} \\
    \midrule
    Berry also announced that \textbf{he} would not be staying on the show on Channel 4, citing "loyalty" to BBC.\\
    \midrule
    \midrule
    \cellcolor[rgb]{0.97, 0.97, 0.97}
    \hspace{6.0em} {\Large \color{colorcorrect} $\bullet$} \textbf{Correct Explanation} \\ 
    \midrule
    No. The summary incorrectly refers to Berry as "he" instead of "she". -- \texttt{GPT-4} \\
    \midrule
    \cellcolor[rgb]{0.97, 0.97, 0.97}
    \hspace{3.5em} {\Large \color{colorpartiallycorrect} $\bullet$} \textbf{Partially Correct Explanation} \\
    \midrule
    The summary incorrectly states that Mary Berry is a he and that Berry cited loyalty to the BBC as a reason for not remaining on the show.  -- \texttt{Claude V1.3}  \\
    \midrule
    \cellcolor[rgb]{0.97, 0.97, 0.97}
    \hspace{6.0em} {\Large \color{colornoexplanation} $\bullet$} \textbf{No Explanation}  \\
    \midrule 
    No -- \texttt{Bab001}  \\
    \midrule
    \cellcolor[rgb]{0.97, 0.97, 0.97}
    \hspace{5.0em} {\Large \color{colorunrelated} $\bullet$} \textbf{Unrelated Explanation} \\
    \midrule
    For me the answer is No! The document says "The new BBC series returns this autumn on Channel 4 after it outbid the BBC". -- \texttt{LlaMa-13b} \\
    \midrule
    \cellcolor[rgb]{0.97, 0.97, 0.97}
    \hspace{5.5em} {\Large \color{colorincorrect} $\bullet$} \textbf{Incorrect Explanation} \\
    \midrule
    The document does not mention Mary Berry announcing that she would not be staying on the show on Channel 4. -- \texttt{Alpaca-13b} \\
    \bottomrule
    \end{tabular}
    }
    \caption{Examples of explanation types used to evaluate model ability at explaining inconsistencies they detect.}
    \label{table:example_explanations}
\end{table}

To gain further insights into LLM ability to reason about factual consistency, we performed a manual analysis of more than 3,600 explanations generated for sixteen of the seventeen LLMs included in our experiments\footnote{PaLM-Bison was released shortly before the publication of this work and could not be included in the manual analysis.}.

In our analysis, we focus on cases the model classifies as inconsistent, as there is a clear explanation a model should provide to pinpoint elements in the summary it identifies as inconsistent. %a clearer signal to look for than explanations justifying a consistent summary.

For each known inconsistent \texttt{(document, summary)} sample in FactCC, and each model output explaining why the sample is inconsistent, we hired an annotator to label the explanation with one of five labels: {\Large \color{colorcorrect} $\bullet$} entirely correct: the model's full explanation must accurately describe a factual inaccuracy in the summary, {\Large \color{colorpartiallycorrect} $\bullet$} partially correct: the model correctly describes at least one factual inconsistency in the summary, but also describes an incorrect element or facts unrelated to factual consistency, {\Large \color{colornoexplanation} $\bullet$} no explanation: the model provides a classification label (Yes/No) without providing the requested explanation, {\Large \color{colorunrelated} $\bullet$} unrelated: the model's output addresses aspects other than factual consistency or is not an explanation (e.g., the model writing a new summary), and {\Large \color{colorincorrect} $\bullet$} incorrect: the model's explanation is invalid and does not correctly identify an element in the summary which is factually incorrect. Table~\ref{table:example_explanations} gives an example of each explanation type from the annotation, and Appendix~\ref{appendix:reasoning_guidelines} provides further information on the hiring and onboarding of the two annotators that completed the task. During annotation, the annotator samples were presented in a shuffled order, and the annotator was not aware of the models that had generated any particular sample.

We analyze annotation reproducibility by collecting multiple annotations for roughly 200 annotations and computing Cohen's Kappa. We find a moderate agreement amongst annotators of 0.72 on the five-way annotation.

\begin{figure}
    \centering
    \includegraphics[width=0.48\textwidth]{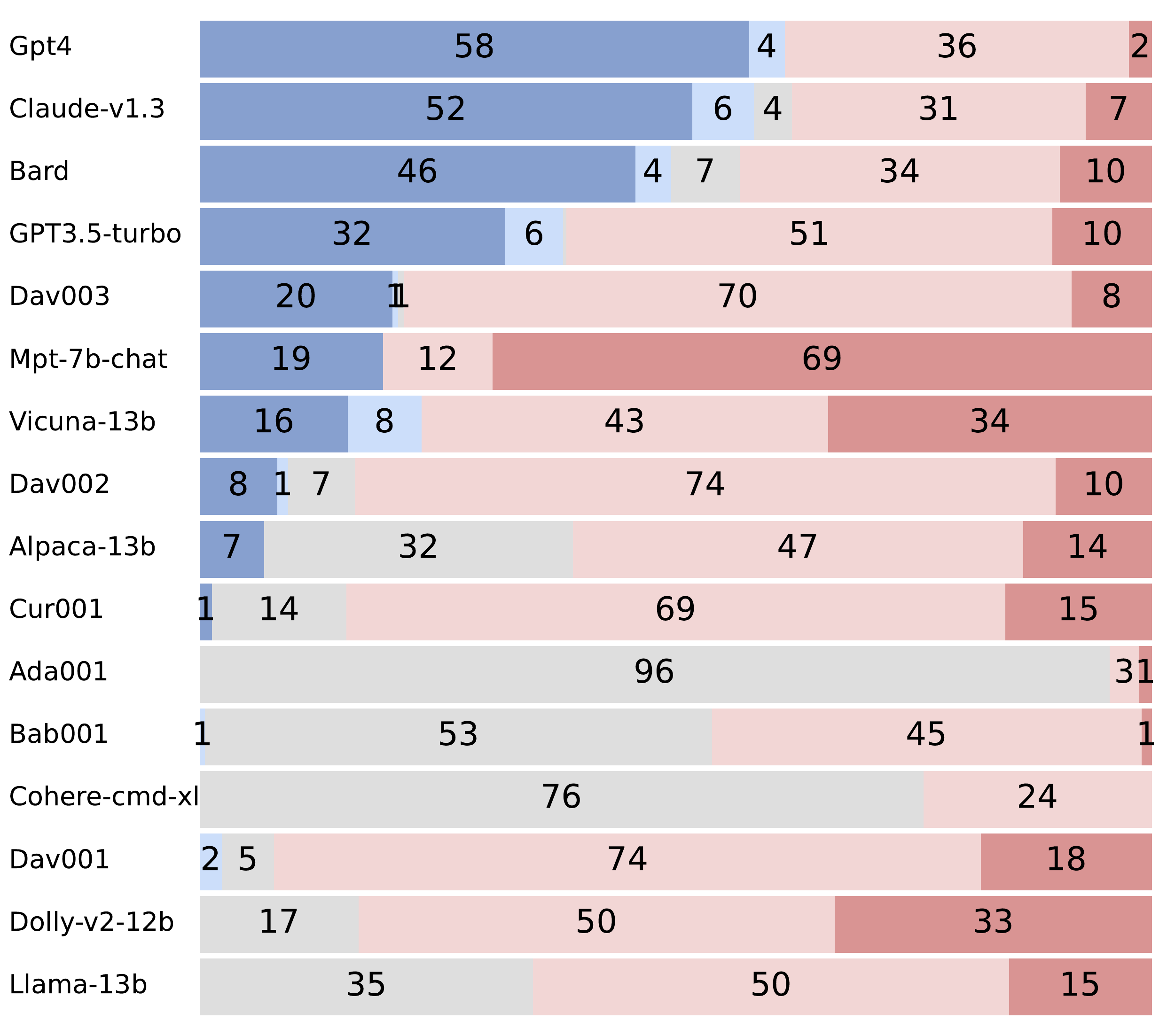}
    \caption{Percentage distribution of the types of explanations each LLM provides in its output when predicting a FactCC summary is inconsistent. Each model explanation is manually annotated as {\Large \color{colorcorrect} $\bullet$} entirely correct, {\Large \color{colorpartiallycorrect} $\bullet$} partially correct, {\Large \color{colornoexplanation} $\bullet$} no explanation provided, {\Large \color{colorunrelated} $\bullet$} unrelated to factuality, {\Large \color{colorincorrect} $\bullet$} or incorrect.}
    \label{fig:factcc_explanation_analysis}
\end{figure}

Figure~\ref{fig:factcc_explanation_analysis} summarizes the results by providing the distribution of types of explanations generated by each LLM. 

First, we find that all models struggle to provide correct explanations pinpointing the inconsistencies in summaries, with nine of the sixteen models providing correct explanations less than 10\% of the time and only three models -- Bard, Claude V1.3, and GPT4 -- providing correct explanations more than 50\% of the time.

We also notice that better performance at the binary classification task (Table~\ref{table:factcc-bacc}) does not necessarily lead to more accurate model explanations. For example, GPT3.5-turbo performs 5-10\% better in terms of binary accuracy than Claude V1.3, yet it generates almost half as many correct explanations. This finding suggests accuracy metrics sometimes overlook models that are \textit{right for the wrong reasons} \cite{mccoy2019right}, which might negatively affect user trust.

% Comparing the outcomes of the inconsistency detection experiments in Table~\ref{table:factcc-bacc} with the factual reasoning abilities of models we notice that the order of evaluators with respect to their performance differs. These results show that models which are good classifiers might not excel at explaining their predictions (GPT3.5-turbo).

Analyzing the failure behavior of models reveals differences amongst models. The first group of models -- including Dav001, Dav002, and Cur001 -- fails by not providing an explanation for the inconsistency, even though they were explicitly prompted to accompany their answer with an explanation. A second group -- including Ada001, LLaMa-13B, and Cohere-cmd-XL -- most often generates ``unrelated explanations'', which do not explain the nature of factual inconsistency, but might instead quote other facts omitted from the summary, propose a new summary, or other tangential text. A third group -- with models such as MPT-7B-Chat and Dolly-v2-12B -- generates plausible explanations that are factually incorrect, or present invalid logic. We argue that although all models should strive to produce only correct explanations, some failure cases are preferable to others. For example, it might be preferable for a model to provide no explanation than a plausible-looking but incorrect one that might mislead a user. For example, MPT-7B-Chat and Dav003 both generate roughly the same proportion of correct explanations (21 vs. 24\%), yet when they fail, Dav003 is much more likely to abstain from providing an explanation, whereas MPT-7B-Chat is more likely to provide an explanation with incorrect reasoning, which could prove more harmful.

\subsection{Fine-grained Inconsistency Detection}
%Previously described prompts were targeted at general factuality evaluation where the goal was to detect any (unspecified) type of factual inconsistency. To explore the model's understanding of factual correctness on a fine-grained level, we leverage the error-type labels assigned to FactCC examples. Fine-grained prompt query the models with evaluating factuality with respect to only one specific type of error, e.g. entity-related errors. Here, we explore individual prompts where each error type has an associated prompt. We also conduct experiments where individual prompts are combined with few-shot examples to aid the model in understanding the task.

To explore the LLMs' fine-grained understanding of factual evaluation we designed a set of experiments prompting the models to evaluate each \texttt{(document, sentence)} pair with respect to individual error types. For each of the error types present in the data models were expected to overlook any other factual error, thus evaluating model ability to selectively focus on one error type while ignoring others. In the experiments, we filter out consistent summaries, and examples containing the error type associated with the prompt were considered "positive", while other error types are considered as negative, and measured performance in terms of Precision and Recall. We use individual prompts for each of the error types and also conduct experiments where individual prompts are combined with few-shot examples to aid the model in understanding the task. Results are presented in Table~\ref{table:factcc-fg-bacc}.

The results show a consistent pattern across all considered LLMs and error types, where the models achieve a low precision and high recall score. This indicates that the models are not able to follow the fine-grained instructions and distinguish different error types. Instead, they simply detect factual inconsistencies (to the best of their abilities) on a general level and assign a negative label. Providing the models with examples of the error to be detected (few-shot setting) does improve the performance of a subset of models; however, no general performance improvement patterns emerged.

In short, none of the models we experiment with can perform the task of fine-grained factual inconsistency detection, in which they are tasked with focusing on a single type of factual error.

Additionally, we carried out an experiment where the per-error-type prompts were combined into a single instruction with multiple tasks that the model was expected to complete in a sequence. Qualitative analysis of the results showed that most of the models could not follow the instructions and consistently provide answers in the expected format, thus the results were not included in the final results table.
\begin{table*}[]
\centering
\resizebox{0.98\textwidth}{!}{%
\begin{tabular}{lcccccccccccccccccccc}
& \multicolumn{10}{c}{Zero-Shot} & \multicolumn{10}{c}{Few-Shot} \\
%\toprule
\cmidrule(r){2-11} \cmidrule(r){12-21}
& \multicolumn{2}{c}{DS} & \multicolumn{2}{c}{ES} & \multicolumn{2}{c}{NSent} & \multicolumn{2}{c}{NS} & \multicolumn{2}{c}{PS} & \multicolumn{2}{c}{DS} & \multicolumn{2}{c}{ES} & \multicolumn{2}{c}{NSent} & \multicolumn{2}{c}{NS} & \multicolumn{2}{c}{PS} \\
\midrule
  Model ($\downarrow$) & P & R & P & R & P & R & P & R & P & R & P & R & P & R & P & R & P & R & P & R \\
\midrule
LLaMa-13B & \cellcolor[rgb]{0.90, 0.72, 0.72} 12.0 & \cellcolor[rgb]{0.90, 0.72, 0.72} 12.0 & \cellcolor[rgb]{0.93, 0.81, 0.81} 18.8 & \cellcolor[rgb]{0.90, 0.72, 0.72} 12.0 & \cellcolor[rgb]{0.98, 0.95, 0.95} 29.4 & \cellcolor[rgb]{0.94, 0.96, 0.99} 40.0 & \cellcolor[rgb]{0.94, 0.84, 0.84} 20.8 & \cellcolor[rgb]{0.94, 0.83, 0.83} 20.0 & \cellcolor[rgb]{0.91, 0.73, 0.73} 12.5 & \cellcolor[rgb]{0.89, 0.67, 0.67} 8.0 & - & - & - & - & - & - & - & - & - & - \\
 Alpaca-13B & \cellcolor[rgb]{0.94, 0.83, 0.83} 20.5 & \cellcolor[rgb]{0.99, 0.98, 0.98} 32.0 & \cellcolor[rgb]{0.91, 0.73, 0.73} 12.9 & \cellcolor[rgb]{0.92, 0.77, 0.77} 16.0 & \cellcolor[rgb]{0.97, 0.98, 1.00} 36.6 & \cellcolor[rgb]{0.76, 0.83, 0.97} 60.0 & \cellcolor[rgb]{0.94, 0.84, 0.84} 20.9 & \cellcolor[rgb]{0.98, 0.98, 1.00} 36.0 & \cellcolor[rgb]{0.95, 0.87, 0.87} 23.1 & \cellcolor[rgb]{0.98, 0.98, 1.00} 36.0 & - & - & - & - & - & - & - & - & - & - \\
 Dolly-v2-12B & \cellcolor[rgb]{0.97, 0.91, 0.91} 26.7 & \cellcolor[rgb]{0.92, 0.77, 0.77} 16.0 & \cellcolor[rgb]{0.93, 0.81, 0.81} 18.8 & \cellcolor[rgb]{0.90, 0.72, 0.72} 12.0 & \cellcolor[rgb]{0.91, 0.75, 0.75} 14.3 & \cellcolor[rgb]{0.92, 0.77, 0.77} 16.0 & \cellcolor[rgb]{0.95, 0.85, 0.85} 22.2 & \cellcolor[rgb]{0.89, 0.67, 0.67} 8.0 & \cellcolor[rgb]{1.00, 1.00, 1.00} 33.3 & \cellcolor[rgb]{0.90, 0.72, 0.72} 12.0 & - & - & - & - & - & - & - & - & - & - \\
 MPT-7B-Chat & \cellcolor[rgb]{0.94, 0.81, 0.81} 19.1 & \cellcolor[rgb]{0.83, 0.88, 0.98} 52.0 & \cellcolor[rgb]{0.94, 0.82, 0.82} 19.7 & \cellcolor[rgb]{0.76, 0.83, 0.97} 60.0 & \cellcolor[rgb]{0.96, 0.87, 0.87} 23.6 & \cellcolor[rgb]{0.69, 0.78, 0.96} 68.0 & \cellcolor[rgb]{0.93, 0.80, 0.80} 18.1 & \cellcolor[rgb]{0.83, 0.88, 0.98} 52.0 & \cellcolor[rgb]{0.93, 0.81, 0.81} 18.8 & \cellcolor[rgb]{0.87, 0.91, 0.98} 48.0 & - & - & - & - & - & - & - & - & - & - \\
 Vicuna-13B & \cellcolor[rgb]{0.94, 0.84, 0.84} 20.8 & \cellcolor[rgb]{0.50, 0.50, 0.78} 100.0 & \cellcolor[rgb]{0.94, 0.81, 0.81} 19.1 & \cellcolor[rgb]{0.59, 0.64, 0.87} 84.0 & \cellcolor[rgb]{0.96, 0.87, 0.87} 23.4 & \cellcolor[rgb]{0.50, 0.50, 0.78} 100.0 & \cellcolor[rgb]{0.95, 0.84, 0.84} 21.4 & \cellcolor[rgb]{0.52, 0.53, 0.80} 96.0 & \cellcolor[rgb]{0.93, 0.80, 0.80} 18.3 & \cellcolor[rgb]{0.62, 0.67, 0.89} 80.0 & - & - & - & - & - & - & - & - & - & - \\
Cohere-CMD-XL &\cellcolor[rgb]{0.95, 0.85, 0.85} 22.1 &\cellcolor[rgb]{0.55, 0.57, 0.82} 92.0 &\cellcolor[rgb]{0.94, 0.82, 0.82} 19.4 &\cellcolor[rgb]{0.59, 0.64, 0.87} 84.0 &\cellcolor[rgb]{0.97, 0.92, 0.92} 26.9 &\cellcolor[rgb]{0.50, 0.50, 0.78} 100.0 &\cellcolor[rgb]{0.95, 0.85, 0.85} 22.1 &\cellcolor[rgb]{0.50, 0.50, 0.78} 100.0 &\cellcolor[rgb]{0.93, 0.79, 0.79} 17.3 &\cellcolor[rgb]{0.64, 0.71, 0.91} 76.0 &\cellcolor[rgb]{0.93, 0.79, 0.79} 17.6 &\cellcolor[rgb]{0.66, 0.74, 0.93} 72.0 &\cellcolor[rgb]{0.95, 0.84, 0.84} 21.2 &\cellcolor[rgb]{0.66, 0.74, 0.93} 72.0 &\cellcolor[rgb]{0.99, 0.97, 0.97} 30.8 &\cellcolor[rgb]{0.52, 0.53, 0.80} 96.0 &\cellcolor[rgb]{0.94, 0.83, 0.83} 20.2 &\cellcolor[rgb]{0.59, 0.64, 0.87} 84.0 &\cellcolor[rgb]{0.94, 0.84, 0.84} 21.1 &\cellcolor[rgb]{0.62, 0.67, 0.89} 80.0 \\
 Claude-v1.3 & \cellcolor[rgb]{0.94, 0.83, 0.83} 20.3 & \cellcolor[rgb]{0.52, 0.53, 0.80} 96.0 & \cellcolor[rgb]{0.94, 0.82, 0.82} 19.7 & \cellcolor[rgb]{0.52, 0.53, 0.80} 96.0 & \cellcolor[rgb]{0.94, 0.82, 0.82} 19.2 & \cellcolor[rgb]{0.55, 0.57, 0.82} 92.0 & \cellcolor[rgb]{0.94, 0.84, 0.84} 20.8 & \cellcolor[rgb]{0.50, 0.50, 0.78} 100.0 & \cellcolor[rgb]{0.94, 0.83, 0.83} 20.3 & \cellcolor[rgb]{0.50, 0.50, 0.78} 100.0 & \cellcolor[rgb]{0.95, 0.85, 0.85} 21.9 & \cellcolor[rgb]{0.55, 0.57, 0.82} 92.0 & \cellcolor[rgb]{0.95, 0.85, 0.85} 22.0 & \cellcolor[rgb]{0.52, 0.53, 0.80} 96.0 & \cellcolor[rgb]{0.95, 0.86, 0.86} 22.9 & \cellcolor[rgb]{0.52, 0.53, 0.80} 96.0 & \cellcolor[rgb]{0.97, 0.90, 0.90} 26.0 & \cellcolor[rgb]{0.50, 0.50, 0.78} 100.0 & \cellcolor[rgb]{0.94, 0.84, 0.84} 21.0 & \cellcolor[rgb]{0.50, 0.50, 0.78} 100.0 \\
PaLM2-Bison & \cellcolor[rgb]{0.94, 0.84, 0.84} 21.1 & \cellcolor[rgb]{0.72, 0.81, 0.97} 64.0 & \cellcolor[rgb]{0.95, 0.84, 0.84} 21.3 & \cellcolor[rgb]{0.64, 0.71, 0.91} 76.0 & \cellcolor[rgb]{0.97, 0.91, 0.91} 26.1 & \cellcolor[rgb]{0.55, 0.57, 0.82} 92.0 & \cellcolor[rgb]{0.96, 0.87, 0.87} 23.6 & \cellcolor[rgb]{0.59, 0.64, 0.87} 84.0 & \cellcolor[rgb]{0.94, 0.83, 0.83} 20.2 & \cellcolor[rgb]{0.62, 0.67, 0.89} 80.0 & \cellcolor[rgb]{0.95, 0.85, 0.85} 21.7 & \cellcolor[rgb]{0.76, 0.83, 0.97} 60.0 & \cellcolor[rgb]{0.93, 0.79, 0.79} 17.5 & \cellcolor[rgb]{0.79, 0.86, 0.98} 56.0 & \cellcolor[rgb]{0.98, 0.94, 0.94} 28.8 & \cellcolor[rgb]{0.59, 0.64, 0.87} 84.0 & \cellcolor[rgb]{0.96, 0.90, 0.90} 25.3 & \cellcolor[rgb]{0.62, 0.67, 0.89} 80.0 & \cellcolor[rgb]{0.94, 0.83, 0.83} 20.2 & \cellcolor[rgb]{0.69, 0.78, 0.96} 68.0 \\
 Ada001 & \cellcolor[rgb]{0.94, 0.82, 0.82} 19.7 & \cellcolor[rgb]{0.55, 0.57, 0.82} 92.0 & \cellcolor[rgb]{0.94, 0.83, 0.83} 20.0 & \cellcolor[rgb]{0.52, 0.53, 0.80} 96.0 & \cellcolor[rgb]{0.94, 0.82, 0.82} 19.3 & \cellcolor[rgb]{0.55, 0.57, 0.82} 92.0 & \cellcolor[rgb]{0.94, 0.82, 0.82} 19.3 & \cellcolor[rgb]{0.55, 0.57, 0.82} 92.0 & \cellcolor[rgb]{0.93, 0.81, 0.81} 18.4 & \cellcolor[rgb]{0.59, 0.64, 0.87} 84.0 & \cellcolor[rgb]{1.00, 1.00, 1.00} 33.3 & \cellcolor[rgb]{0.87, 0.62, 0.62} 4.0 & \cellcolor[rgb]{0.85, 0.56, 0.56} 0.0 & \cellcolor[rgb]{0.85, 0.56, 0.56} 0.0 & \cellcolor[rgb]{0.85, 0.56, 0.56} 0.0 & \cellcolor[rgb]{0.85, 0.56, 0.56} 0.0 & \cellcolor[rgb]{0.95, 0.86, 0.86} 22.8 & \cellcolor[rgb]{0.55, 0.57, 0.82} 92.0 & \cellcolor[rgb]{0.94, 0.83, 0.83} 20.0 & \cellcolor[rgb]{0.87, 0.62, 0.62} 4.0 \\
 Bab001 & \cellcolor[rgb]{0.87, 0.63, 0.63} 4.8 & \cellcolor[rgb]{0.87, 0.62, 0.62} 4.0 & \cellcolor[rgb]{0.94, 0.81, 0.81} 19.0 & \cellcolor[rgb]{0.92, 0.77, 0.77} 16.0 & \cellcolor[rgb]{0.93, 0.80, 0.80} 18.2 & \cellcolor[rgb]{0.96, 0.88, 0.88} 24.0 & \cellcolor[rgb]{0.90, 0.70, 0.70} 10.3 & \cellcolor[rgb]{0.90, 0.72, 0.72} 12.0 & \cellcolor[rgb]{0.90, 0.72, 0.72} 11.5 & \cellcolor[rgb]{0.90, 0.72, 0.72} 12.0 & \cellcolor[rgb]{0.97, 0.92, 0.92} 27.3 & \cellcolor[rgb]{0.96, 0.88, 0.88} 24.0 & \cellcolor[rgb]{0.94, 0.83, 0.83} 20.0 & \cellcolor[rgb]{0.96, 0.88, 0.88} 24.0 & \cellcolor[rgb]{0.96, 0.87, 0.87} 23.5 & \cellcolor[rgb]{0.99, 0.98, 0.98} 32.0 & \cellcolor[rgb]{0.91, 0.73, 0.73} 12.5 & \cellcolor[rgb]{0.92, 0.77, 0.77} 16.0 & \cellcolor[rgb]{0.94, 0.82, 0.82} 19.4 & \cellcolor[rgb]{0.98, 0.93, 0.93} 28.0 \\
 Cur001 & \cellcolor[rgb]{0.91, 0.74, 0.74} 13.5 & \cellcolor[rgb]{0.98, 0.93, 0.93} 28.0 & \cellcolor[rgb]{0.95, 0.85, 0.85} 22.0 & \cellcolor[rgb]{0.90, 0.93, 0.99} 44.0 & \cellcolor[rgb]{1.00, 0.99, 0.99} 32.8 & \cellcolor[rgb]{0.59, 0.64, 0.87} 84.0 & \cellcolor[rgb]{0.93, 0.80, 0.80} 17.9 & \cellcolor[rgb]{0.98, 0.93, 0.93} 28.0 & \cellcolor[rgb]{0.92, 0.77, 0.77} 15.4 & \cellcolor[rgb]{0.96, 0.88, 0.88} 24.0 & \cellcolor[rgb]{0.92, 0.78, 0.78} 16.4 & \cellcolor[rgb]{0.90, 0.93, 0.99} 44.0 & \cellcolor[rgb]{0.92, 0.77, 0.77} 15.9 & \cellcolor[rgb]{0.90, 0.93, 0.99} 44.0 & \cellcolor[rgb]{0.97, 0.91, 0.91} 26.6 & \cellcolor[rgb]{0.59, 0.64, 0.87} 84.0 & \cellcolor[rgb]{0.91, 0.75, 0.75} 14.3 & \cellcolor[rgb]{0.99, 0.98, 0.98} 32.0 & \cellcolor[rgb]{0.92, 0.78, 0.78} 16.5 & \cellcolor[rgb]{0.83, 0.88, 0.98} 52.0 \\
 Dav001 & \cellcolor[rgb]{0.91, 0.74, 0.74} 13.7 & \cellcolor[rgb]{0.98, 0.93, 0.93} 28.0 & \cellcolor[rgb]{0.97, 0.91, 0.91} 26.3 & \cellcolor[rgb]{0.76, 0.83, 0.97} 60.0 & \cellcolor[rgb]{0.94, 0.96, 0.99} 39.5 & \cellcolor[rgb]{0.69, 0.78, 0.96} 68.0 & \cellcolor[rgb]{0.91, 0.75, 0.75} 14.3 & \cellcolor[rgb]{0.98, 0.93, 0.93} 28.0 & \cellcolor[rgb]{0.91, 0.73, 0.73} 13.0 & \cellcolor[rgb]{0.90, 0.72, 0.72} 12.0 & \cellcolor[rgb]{0.93, 0.81, 0.81} 18.6 & \cellcolor[rgb]{0.83, 0.88, 0.98} 52.0 & \cellcolor[rgb]{0.93, 0.81, 0.81} 18.8 & \cellcolor[rgb]{0.98, 0.98, 1.00} 36.0 & \cellcolor[rgb]{0.96, 0.97, 1.00} 37.5 & \cellcolor[rgb]{0.66, 0.74, 0.93} 72.0 & \cellcolor[rgb]{0.93, 0.79, 0.79} 17.3 & \cellcolor[rgb]{0.98, 0.98, 1.00} 36.0 & \cellcolor[rgb]{0.90, 0.71, 0.71} 10.9 & \cellcolor[rgb]{0.96, 0.88, 0.88} 24.0 \\
 Dav002 & \cellcolor[rgb]{0.94, 0.83, 0.83} 20.2 & \cellcolor[rgb]{0.55, 0.57, 0.82} 92.0 & \cellcolor[rgb]{0.95, 0.86, 0.86} 22.4 & \cellcolor[rgb]{0.52, 0.53, 0.80} 96.0 & \cellcolor[rgb]{0.96, 0.87, 0.87} 23.4 & \cellcolor[rgb]{0.57, 0.60, 0.84} 88.0 & \cellcolor[rgb]{0.95, 0.85, 0.85} 21.9 & \cellcolor[rgb]{0.50, 0.50, 0.78} 100.0 & \cellcolor[rgb]{0.94, 0.83, 0.83} 20.5 & \cellcolor[rgb]{0.55, 0.57, 0.82} 92.0 & \cellcolor[rgb]{0.93, 0.80, 0.80} 18.3 & \cellcolor[rgb]{0.59, 0.64, 0.87} 84.0 & \cellcolor[rgb]{0.94, 0.82, 0.82} 19.8 & \cellcolor[rgb]{0.55, 0.57, 0.82} 92.0 & \cellcolor[rgb]{0.94, 0.84, 0.84} 21.1 & \cellcolor[rgb]{0.52, 0.53, 0.80} 96.0 & \cellcolor[rgb]{0.94, 0.83, 0.83} 20.3 & \cellcolor[rgb]{0.50, 0.50, 0.78} 100.0 & \cellcolor[rgb]{0.94, 0.82, 0.82} 19.7 & \cellcolor[rgb]{0.52, 0.53, 0.80} 96.0 \\
 Dav003 & \cellcolor[rgb]{0.94, 0.83, 0.83} 20.4 & \cellcolor[rgb]{0.55, 0.57, 0.82} 92.0 & \cellcolor[rgb]{0.94, 0.83, 0.83} 20.2 & \cellcolor[rgb]{0.52, 0.53, 0.80} 96.0 & \cellcolor[rgb]{0.96, 0.89, 0.89} 24.7 & \cellcolor[rgb]{0.52, 0.53, 0.80} 96.0 & \cellcolor[rgb]{0.95, 0.85, 0.85} 21.7 & \cellcolor[rgb]{0.50, 0.50, 0.78} 100.0 & \cellcolor[rgb]{0.94, 0.83, 0.83} 20.2 & \cellcolor[rgb]{0.52, 0.53, 0.80} 96.0 & \cellcolor[rgb]{0.97, 0.90, 0.90} 25.6 & \cellcolor[rgb]{0.55, 0.57, 0.82} 92.0 & \cellcolor[rgb]{0.95, 0.85, 0.85} 21.6 & \cellcolor[rgb]{0.52, 0.53, 0.80} 96.0 & \cellcolor[rgb]{0.96, 0.88, 0.88} 24.0 & \cellcolor[rgb]{0.55, 0.57, 0.82} 92.0 & \cellcolor[rgb]{0.97, 0.92, 0.92} 27.5 & \cellcolor[rgb]{0.50, 0.50, 0.78} 100.0 & \cellcolor[rgb]{0.94, 0.82, 0.82} 19.3 & \cellcolor[rgb]{0.55, 0.57, 0.82} 92.0 \\
 GPT3.5-turbo & \cellcolor[rgb]{0.94, 0.83, 0.83} 20.6 & \cellcolor[rgb]{0.57, 0.60, 0.84} 88.0 & \cellcolor[rgb]{0.93, 0.80, 0.80} 18.0 & \cellcolor[rgb]{0.62, 0.67, 0.89} 80.0 & \cellcolor[rgb]{0.96, 0.88, 0.88} 24.0 & \cellcolor[rgb]{0.50, 0.50, 0.78} 100.0 & \cellcolor[rgb]{0.95, 0.86, 0.86} 22.9 & \cellcolor[rgb]{0.50, 0.50, 0.78} 100.0 & \cellcolor[rgb]{0.94, 0.84, 0.84} 21.1 & \cellcolor[rgb]{0.52, 0.53, 0.80} 96.0 & \cellcolor[rgb]{0.94, 0.83, 0.83} 20.2 & \cellcolor[rgb]{0.55, 0.57, 0.82} 92.0 & \cellcolor[rgb]{0.93, 0.78, 0.78} 16.7 & \cellcolor[rgb]{0.64, 0.71, 0.91} 76.0 & \cellcolor[rgb]{0.95, 0.85, 0.85} 21.8 & \cellcolor[rgb]{0.52, 0.53, 0.80} 96.0 & \cellcolor[rgb]{0.95, 0.86, 0.86} 22.7 & \cellcolor[rgb]{0.50, 0.50, 0.78} 100.0 & \cellcolor[rgb]{0.94, 0.84, 0.84} 20.7 & \cellcolor[rgb]{0.52, 0.53, 0.80} 96.0 \\
 GPT4 & \cellcolor[rgb]{0.99, 0.97, 0.97} 31.0 & \cellcolor[rgb]{0.57, 0.60, 0.84} 88.0 & \cellcolor[rgb]{0.94, 0.83, 0.83} 20.5 & \cellcolor[rgb]{0.52, 0.53, 0.80} 96.0 & \cellcolor[rgb]{0.95, 0.85, 0.85} 22.0 & \cellcolor[rgb]{0.52, 0.53, 0.80} 96.0 & \cellcolor[rgb]{0.96, 0.88, 0.88} 24.3 & \cellcolor[rgb]{0.50, 0.50, 0.78} 100.0 & \cellcolor[rgb]{0.94, 0.82, 0.82} 19.8 & \cellcolor[rgb]{0.52, 0.53, 0.80} 96.0 & \cellcolor[rgb]{0.97, 0.91, 0.91} 26.4 & \cellcolor[rgb]{0.55, 0.57, 0.82} 92.0 & \cellcolor[rgb]{0.94, 0.83, 0.83} 20.2 & \cellcolor[rgb]{0.52, 0.53, 0.80} 96.0 & \cellcolor[rgb]{0.95, 0.86, 0.86} 22.4 & \cellcolor[rgb]{0.52, 0.53, 0.80} 96.0 & \cellcolor[rgb]{0.95, 0.85, 0.85} 21.9 & \cellcolor[rgb]{0.50, 0.50, 0.78} 100.0 & \cellcolor[rgb]{0.94, 0.83, 0.83} 20.5 & \cellcolor[rgb]{0.52, 0.53, 0.80} 96.0 \\
\bottomrule
\end{tabular}
}
\caption{Precision (P) and Recall (R) scores of error detection with fine-grained prompts for individual error types. Experiments run in Zero- and Few-shot settings for each of the error types: Date Swap (\texttt{DS}), Entity Swap (\texttt{ES}), Negated Sentences (\texttt{NSent}), Number Swap (\texttt{NS}), Pronoun Swap (\texttt{PS}).}
\label{table:factcc-fg-bacc}
\end{table*}

\section{Limits of Crowd-Based Benchmarks}
In this section we analyze two popular benchmarks for factual consistency detection in summarization: AggreFact \cite{tang2022understanding} and DialSummEval \cite{gao2022dialsummeval} and uncover limitations that guide the design principles of the \dataset{} benchmark we build.

\subsection{Experimental Setup}
\label{section:experimental_setup}

In the subsequent sections of the paper, we incorporate the insights gained from the experiments on FactCC to inform our experiment design in terms of model and prompt selection.

First, we filter out all models that did not achieve a balanced accuracy above 60\% on FactCC, as such models are unlikely to significantly outperform random chance on more challenging benchmarks. Checkmarks (\checkmark) in Table~\ref{table:factcc-bacc} indicate models that are retained in the experiments of Sections~\ref{section:existing_benchmarks}-\ref{section:summedits_benchmark}.

Second, to minimize the computational cost of experiments, we select a single Zero-Shot (ZS) prompt that is used for all LLM models. We make this choice instead of using multiple prompts per model or selecting each model's best-performing prompt on FactCC results for three reasons: (1) there's no guarantee that prompt quality will transfer across benchmarks, and using a single common prompt removes variance from prompt optimization that does not measure underlying model ability, (2) top-performing LLMs such as GPT4 achieve their best performance on FactCC with ZS prompts, indicating that high performance with a simple ZS prompt is achievable, and (3) more complex prompts would require adaptation to each domain (e.g. domain-specific few-shot examples), and restrict the evaluation of models with shorter maximum sequence lengths due to longer prompts.

\label{section:existing_benchmarks}
\begin{table}[]
    \centering
    \resizebox{0.4\textwidth}{!}{%
    \begin{tabular}{lccc}
     & \textbf{AggreFact} & \multicolumn{2}{c}{\textbf{DialSummEval}} \\
    \cmidrule(r){1-1} \cmidrule(r){2-2} \cmidrule(r){3-4}
    \textbf{Model Name} & \textbf{\%BAcc.} & \textbf{\%BAcc.} & \textbf{Corr.} \\
    \cmidrule(r){1-1} \cmidrule(r){2-2} \cmidrule(r){3-4}
     DAE & \cellcolor[rgb]{0.64, 0.71, 0.91} \textbf{76.0} & \cellcolor[rgb]{0.91, 0.73, 0.73} 56.2 & \cellcolor[rgb]{0.90, 0.93, 0.99} 0.44 \\
    SummaC & \cellcolor[rgb]{0.92, 0.94, 0.99} 71.6 & \cellcolor[rgb]{0.96, 0.89, 0.89} 62.7 & \cellcolor[rgb]{0.96, 0.96, 1.00} 0.35 \\
    QAFactEval & \cellcolor[rgb]{0.86, 0.91, 0.98} 73.9 & \cellcolor[rgb]{0.98, 0.94, 0.94} 64.4 & \cellcolor[rgb]{0.77, 0.84, 0.97} \textbf{0.59} \\
    \cmidrule(r){1-1} \cmidrule(r){2-2} \cmidrule(r){3-4}
    Cohere-cmd-XL & \cellcolor[rgb]{0.96, 0.89, 0.89} 63.1 & \cellcolor[rgb]{0.91, 0.73, 0.73} 56.6 & \cellcolor[rgb]{0.96, 0.96, 1.00} 0.36 \\
    Claude V1.3 & \cellcolor[rgb]{0.86, 0.59, 0.59} 50.6 & \cellcolor[rgb]{0.91, 0.73, 0.73} 56.8 & \cellcolor[rgb]{0.99, 0.97, 0.97} 0.30 \\
    Bard & \cellcolor[rgb]{0.96, 0.89, 0.89} 62.7 & \cellcolor[rgb]{0.93, 0.81, 0.81} 59.5 & \cellcolor[rgb]{0.96, 0.89, 0.89} 0.26 \\
    PaLM2-Bison    & \cellcolor[rgb]{0.91, 0.75, 0.75} 57.0 & \cellcolor[rgb]{0.90, 0.70, 0.70} 55.6 & 0.57 \\
    Dav001 & \cellcolor[rgb]{0.88, 0.65, 0.65} 53.3 & \cellcolor[rgb]{0.88, 0.65, 0.65} 52.9 & \cellcolor[rgb]{0.90, 0.71, 0.71} 0.11 \\
    Dav002 & \cellcolor[rgb]{0.89, 0.67, 0.67} 54.3 & \cellcolor[rgb]{0.93, 0.81, 0.81} 59.2 & \cellcolor[rgb]{0.87, 0.91, 0.99} 0.49 \\
    Vicuna-13b &  \cellcolor[rgb]{0.95, 0.84, 0.84} 60.3 & \cellcolor[rgb]{0.92, 0.78, 0.78} 58.6 & \cellcolor[rgb]{0.96, 0.96, 1.00} 0.36 \\
    Dav003 & \cellcolor[rgb]{0.98, 0.95, 0.95} 64.8 & \cellcolor[rgb]{0.95, 0.84, 0.84} 60.9 & \cellcolor[rgb]{0.81, 0.87, 0.98} 0.51 \\
    GPT3.5-turbo & \cellcolor[rgb]{0.94, 0.96, 0.99} 70.2 & \cellcolor[rgb]{0.96, 0.89, 0.89} 62.0 & \cellcolor[rgb]{0.78, 0.85, 0.98} 0.56 \\
    GPT-4 & \cellcolor[rgb]{0.86, 0.91, 0.98} 73.6 & \cellcolor[rgb]{0.97, 0.98, 1.00} \textbf{68.4} & \cellcolor[rgb]{0.77, 0.84, 0.97} 0.58 \\
    \bottomrule{}
    \end{tabular}
    }
    \caption{Performance of models on the AggreFact, DialSummEval consistency benchmarks reported in balanced accuracy (\textbf{\%Bacc.}) and correlation (\textbf{corr.}).}
    \label{table:existing_benchmarks}
\end{table}

\subsection{AggreFact}
% \begin{figure}
%     \centering
%     \includegraphics[width=0.49\textwidth]{figures/AggreFact_GPT4_Invalidation.pdf}
%     \caption{Manual analysis of all samples in AggreFact that GPT4 predicts as inconsistent. For 74 of the 101 samples marked as consistent in the dataset (\texttt{label = 1)}, GPT4's explanation is correct: GPT4 has correctly identified an inconsistency, invalidating the dataset's label.}
%     \label{fig:aggrefact_explanations}
% \end{figure}

The AggreFact-SOTA \cite{tang2022understanding} benchmark is a factual consistency benchmark focused on the news domain, modified from the SummaC benchmark \cite{laban2022summac} focused on summaries generated by \textit{SOTA} models (i.e., models based on pre-trained Transformers), as analysis showed that summaries from older models were less relevant to the field of consistency detection.

Table~\ref{fig:benchmark_comparison} reports the balanced accuracy of specialized models and LLMs on AggreFact. At first glance, the specialized models still outperform LLMs, even though increasing LLM size leads to performance improvements and helps close the gap, with GPT-4 performing within 2.4\% points of the specialized DAE. However, all models perform relatively poorly, with no model reaching a balanced accuracy of 80\% on a binary classification task.

To inspect performance on the AggreFact benchmark further, we conducted a manual annotation similar to the one conducted in FactCC Section~\ref{section:factcc_reasoning} but focused on cases where GPT4 disagrees with the label of AggreFact. More precisely, we manually inspected the explanations provided by GPT4 for the 101 summaries it judged were inconsistent but labeled as consistent in the dataset.

Of the 101 samples, 80 were labeled by the annotator as correct or partially correct explanations that identify and explain a factual inconsistency in the summary. In other words, this manual analysis of a subset of AggreFact reveals that \textbf{a minimum of 6\% of the samples in AggreFact are mislabeled.} The low reliability of labels in crowd-sourced benchmarks like AggreFact is a known issue \cite{pagnoni2021understanding} stemming from task complexity that requires the annotator to carefully read and understand an entire document and accompanying summary, leading to low repeatability and inter-annotator agreement.

This methodology reveals the potential for LLMs as part of dataset creation. In some cases, an LLM explanation that is verifiable -- such as an explanation for an identified factual inconsistency -- can accelerate and improve the quality of annotation. We note however that LLM explanations are only valuable for a subset of the samples. For example, in cases where the model asserts a summary is consistent, manual verification is still required to assure quality. In Section~\ref{section:summedits_benchmark}, we explore a new protocol for factual consistency benchmark creation which can involve an LLM.

Based on the low reliability of labels in AggreFact, we note that a key requirement for future benchmarks is to improve label reliability, which can be demonstrated with high annotator agreement when multiple annotators are involved.

\subsection{DialSummEval}
\begin{table}[]
    \centering
\renewcommand{\arraystretch}{1.2} 
\resizebox{0.48\textwidth}{!}{%
\begin{tabular}{lcc|cccc|cc}
 & \multicolumn{8}{c}{\textbf{Average Annotator Likert Score}} \\
\cmidrule{2-9}
\textbf{Model} & \cellcolor[rgb]{0.93, 0.79, 0.79} 1.5 & \cellcolor[rgb]{0.93, 0.79, 0.79} 2.0 & \cellcolor[rgb]{0.87, 0.87, 0.87} 2.5 & \cellcolor[rgb]{0.87, 0.87, 0.87} 3.0 & \cellcolor[rgb]{0.87, 0.87, 0.87} 3.5 & \cellcolor[rgb]{0.87, 0.87, 0.87} 4.0 & \cellcolor[rgb]{0.61, 0.66, 0.88} 4.5 & \cellcolor[rgb]{0.61, 0.66, 0.88} 5.0 \\
\cmidrule(r){1-1} \cmidrule{2-9}
Dav001     &       \cellcolor[rgb]{0.69, 0.78, 0.96} 68.1 &       \cellcolor[rgb]{0.63, 0.69, 0.90} 78.4 &       \cellcolor[rgb]{0.59, 0.63, 0.86} 84.6 &       \cellcolor[rgb]{0.56, 0.59, 0.83} 90.2 &       \cellcolor[rgb]{0.60, 0.64, 0.87} 83.6 &       \cellcolor[rgb]{0.59, 0.63, 0.86} 84.9 &       \cellcolor[rgb]{0.58, 0.62, 0.85} 86.0 &       \cellcolor[rgb]{0.56, 0.60, 0.84} 88.9 \\
Cohere-cmd-XL &\cellcolor[rgb]{0.88, 0.92, 0.99} 46.2 &\cellcolor[rgb]{0.84, 0.89, 0.98} 51.0 &\cellcolor[rgb]{0.67, 0.76, 0.94} 70.3 &\cellcolor[rgb]{0.60, 0.64, 0.87} 83.6 &\cellcolor[rgb]{0.57, 0.60, 0.84} 88.6 &\cellcolor[rgb]{0.56, 0.59, 0.84} 89.2 &\cellcolor[rgb]{0.55, 0.57, 0.82} 91.7 &\cellcolor[rgb]{0.52, 0.53, 0.80} 96.3 \\
DAE        &       \cellcolor[rgb]{0.99, 0.97, 0.97} 30.8 &       \cellcolor[rgb]{0.78, 0.85, 0.98} 56.9 &       \cellcolor[rgb]{0.72, 0.81, 0.97} 63.7 &       \cellcolor[rgb]{0.60, 0.64, 0.87} 83.6 &       \cellcolor[rgb]{0.58, 0.61, 0.85} 86.8 &       \cellcolor[rgb]{0.53, 0.55, 0.81} 94.3 &       \cellcolor[rgb]{0.56, 0.58, 0.83} 90.3 &       \cellcolor[rgb]{0.53, 0.55, 0.81} 94.2 \\
PaLM2-bison &\cellcolor[rgb]{0.96, 0.90, 0.90} 25.3 &\cellcolor[rgb]{0.98, 0.99, 1.00} 35.3 &\cellcolor[rgb]{0.79, 0.86, 0.98} 56.0 &\cellcolor[rgb]{0.62, 0.69, 0.90} 78.7 &\cellcolor[rgb]{0.54, 0.56, 0.81} 93.6 &\cellcolor[rgb]{0.52, 0.52, 0.79} 97.2 &\cellcolor[rgb]{0.51, 0.51, 0.78} 98.4 &\cellcolor[rgb]{0.52, 0.54, 0.80} 95.8 \\
Dav002     &       \cellcolor[rgb]{0.91, 0.74, 0.74} 13.2 &       \cellcolor[rgb]{0.98, 0.95, 0.95} 29.4 &       \cellcolor[rgb]{0.87, 0.91, 0.99} 47.3 &       \cellcolor[rgb]{0.73, 0.82, 0.97} 62.3 &       \cellcolor[rgb]{0.63, 0.69, 0.90} 77.7 &       \cellcolor[rgb]{0.60, 0.65, 0.87} 83.0 &       \cellcolor[rgb]{0.57, 0.60, 0.84} 88.3 &       \cellcolor[rgb]{0.56, 0.59, 0.83} 90.0 \\
Dav003     &        \cellcolor[rgb]{0.87, 0.62, 0.62} 4.4 &       \cellcolor[rgb]{0.93, 0.79, 0.79} 17.6 &       \cellcolor[rgb]{0.98, 0.94, 0.94} 28.6 &       \cellcolor[rgb]{0.99, 0.97, 0.97} 31.1 &       \cellcolor[rgb]{0.73, 0.81, 0.97} 63.2 &       \cellcolor[rgb]{0.68, 0.77, 0.95} 69.3 &       \cellcolor[rgb]{0.59, 0.63, 0.86} 84.9 &       \cellcolor[rgb]{0.61, 0.66, 0.88} 81.6 \\
GPT3.5-turbo       &        \cellcolor[rgb]{0.89, 0.68, 0.68} 8.8 &       \cellcolor[rgb]{0.92, 0.77, 0.77} 15.7 &       \cellcolor[rgb]{0.98, 0.95, 0.95} 29.7 &       \cellcolor[rgb]{0.89, 0.92, 0.99} 45.9 &       \cellcolor[rgb]{0.65, 0.73, 0.93} 73.6 &       \cellcolor[rgb]{0.64, 0.71, 0.91} 76.4 &       \cellcolor[rgb]{0.57, 0.60, 0.84} 88.5 &       \cellcolor[rgb]{0.56, 0.59, 0.83} 90.0 \\
GPT4       &        \cellcolor[rgb]{0.86, 0.59, 0.59} 2.2 &        \cellcolor[rgb]{0.88, 0.64, 0.64} 5.9 &        \cellcolor[rgb]{0.88, 0.65, 0.65} 6.6 &       \cellcolor[rgb]{0.96, 0.89, 0.89} 24.6 &       \cellcolor[rgb]{0.89, 0.92, 0.99} 45.9 &       \cellcolor[rgb]{0.81, 0.87, 0.98} 54.2 &       \cellcolor[rgb]{0.61, 0.67, 0.88} 80.9 &       \cellcolor[rgb]{0.57, 0.61, 0.84} 87.9 \\
QAFactEval &        \cellcolor[rgb]{0.86, 0.61, 0.61} 3.3 &        \cellcolor[rgb]{0.88, 0.64, 0.64} 5.9 &       \cellcolor[rgb]{0.93, 0.79, 0.79} 17.6 &       \cellcolor[rgb]{0.96, 0.89, 0.89} 24.6 &       \cellcolor[rgb]{0.90, 0.93, 0.99} 44.5 &       \cellcolor[rgb]{0.80, 0.87, 0.98} 54.7 &       \cellcolor[rgb]{0.67, 0.76, 0.94} 70.3 &       \cellcolor[rgb]{0.65, 0.72, 0.92} 74.7 \\
Vicuna-13b &\cellcolor[rgb]{0.89, 0.68, 0.68} 8.8 &\cellcolor[rgb]{0.92, 0.77, 0.77} 15.7 &\cellcolor[rgb]{0.93, 0.79, 0.79} 17.6 &\cellcolor[rgb]{0.96, 0.97, 1.00} 37.7 &\cellcolor[rgb]{0.84, 0.89, 0.98} 50.9 &\cellcolor[rgb]{0.81, 0.87, 0.98} 54.2 &\cellcolor[rgb]{0.71, 0.80, 0.97} 65.5 &\cellcolor[rgb]{0.69, 0.79, 0.96} 66.8 \\
SummaC &        \cellcolor[rgb]{0.87, 0.62, 0.62} 4.4 &        \cellcolor[rgb]{0.88, 0.64, 0.64} 5.9 &       \cellcolor[rgb]{0.94, 0.84, 0.84} 20.9 &       \cellcolor[rgb]{0.95, 0.84, 0.84} 21.3 &       \cellcolor[rgb]{0.97, 0.93, 0.93} 27.7 &       \cellcolor[rgb]{0.94, 0.96, 0.99} 40.1 &       \cellcolor[rgb]{0.91, 0.93, 0.99} 43.7 &       \cellcolor[rgb]{0.77, 0.84, 0.97} 58.9 \\
Claude V1.3 &        \cellcolor[rgb]{0.85, 0.58, 0.58} 1.1 &        \cellcolor[rgb]{0.89, 0.69, 0.69} 9.8 &       \cellcolor[rgb]{0.90, 0.71, 0.71} 11.0 &       \cellcolor[rgb]{0.91, 0.74, 0.74} 13.1 &       \cellcolor[rgb]{1.00, 1.00, 1.00} 33.6 &       \cellcolor[rgb]{0.96, 0.98, 1.00} 37.3 &       \cellcolor[rgb]{0.87, 0.91, 0.99} 47.1 &       \cellcolor[rgb]{0.89, 0.92, 0.99} 45.8 \\
Bard       &        \cellcolor[rgb]{0.89, 0.69, 0.69} 9.9 &        \cellcolor[rgb]{0.89, 0.67, 0.67} 7.8 &        \cellcolor[rgb]{0.87, 0.64, 0.64} 5.5 &        \cellcolor[rgb]{0.89, 0.69, 0.69} 9.8 &       \cellcolor[rgb]{0.93, 0.80, 0.80} 18.2 &       \cellcolor[rgb]{0.95, 0.84, 0.84} 21.2 &       \cellcolor[rgb]{0.97, 0.98, 1.00} 36.5 &       \cellcolor[rgb]{0.92, 0.94, 0.99} 42.6 \\
\bottomrule
\end{tabular}
}
    \caption{Percent of summaries classified as consistent in DialSummEval, bucketed by average Likert consistency score. Models are more uncertain in mid-range borderline buckets ([2.0, 4.0]).}
    \label{fig:dialsummeval_detail}
\end{table}

The DialSummEval \cite{gao2022dialsummeval} benchmark is a summarization evaluation benchmark created following the format of SummEval \cite{fabbri2021summeval} for the domain of dialogue summarization. In DialSummEval, each \texttt{(dialogue, summary)} tuple is evaluated by three annotators, each assigning a Likert score (1-5) assessing the consistency of the summary. The authors of the benchmark report an agreement level of 0.67 Krippendorff's alpha on the labels, indicating a moderate amount of agreement among annotators.

We evaluate model performance in two ways: (1) direct correlation between model predictions and the average annotator score, and (2) we follow \citet{laban2022summac}'s procedure to transform the benchmark into a binary classification task, amenable to the balanced accuracy metric. Results are summarized in Table~\ref{table:existing_benchmarks}.

Echoing results on AggreFact, increasing model size leads to a minor improvement in performance both in balanced accuracy and correlation, but most LLMs still underperform specialized methods. In absolute terms, all methods struggle to achieve strong performance on the benchmark, with accuracies all below 70\%.

In Figure~\ref{fig:dialsummeval_detail}, we aggregate model predictions into 0.5-width buckets on the Likert scale. We find that most models achieve strong performance on non-borderline buckets ([1.0, 1.5), [1.5, 2.0], [4.0, 4.5], [4.5, 5.0]), assigning a vast majority of samples to the correct class (inconsistent for low buckets, consistent for high buckets). The borderline buckets ([2.0, 4.0]) however are less clear-cut: most models assign large proportions of samples from each bucket into consistent and inconsistent classes.

We argue that \textbf{annotating the consistency of summaries using a Likert scale limits the quality and interpretability of the benchmark}, as it is not evident to interpret the differences between scores, limiting reproducibility, which is reflected in the moderate Kripendorff's alpha. Instead, we favor framing factual consistency benchmarks as a \textit{detection task}. In the detection task, identifying any factual inconsistency between the document and summary leads to an overall assessment of the summary being \textit{inconsistent}. If no inconsistency is detected, the summary is \textit{consistent}. The detection framing also allows for models to provide natural language explanations when identifying a summary as inconsistent, which can be manually verified to confirm model reasoning ability, and model failure modes, as done in Section~\ref{section:factcc_reasoning}.

In the next section, we propose a novel protocol to create factual consistency benchmarks, incorporating lessons learned from existing benchmarks.

\section{\dataset{} Protocol}
\label{section:protocol}
\begin{figure}
    \centering
    \includegraphics[width=0.48\textwidth]{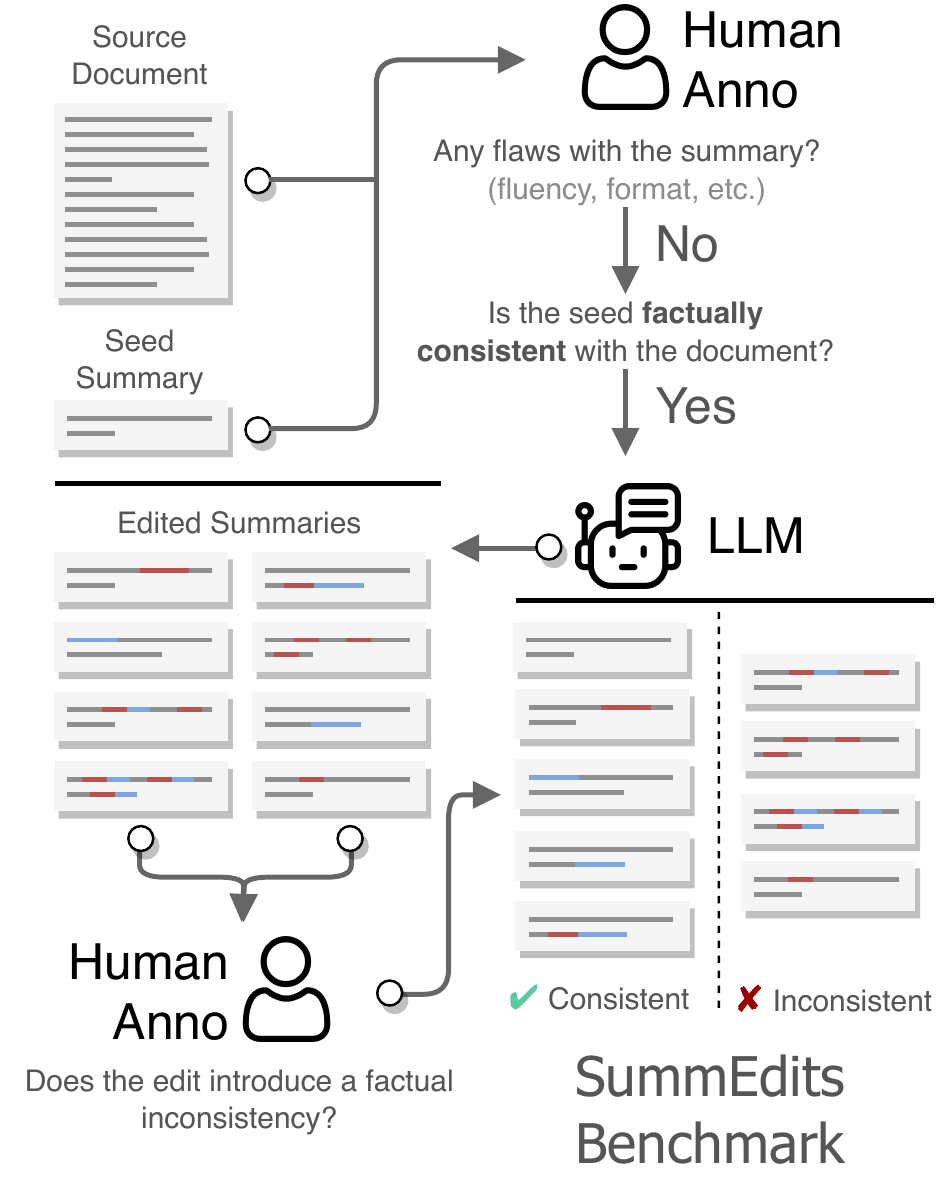}
    \caption{\dataset{} protocol diagram, a three-step protocol to create summarization ID benchmarks. See Table~\ref{table:example_edits} for example samples produced by the protocol.}
    \label{fig:summedits_protocol}
\end{figure}

\begin{table*}[]
    \centering
    \resizebox{0.8\textwidth}{!}{%
    \begin{tabular}{p{8cm} p{8cm}}
    \toprule
    \hspace{1.7em} Edited Summary Labeled As \textbf{Consistent} & \hspace{1.5em} Edited Summary Labeled As \textbf{Inconsistent} \\ 
    \midrule
    
    The characters \hlred{discuss} \hlblue{ponder} the consequences of banishing Marcius, with Cominius warning that his \hlred{alliance} \hlblue{collaboration} with the Volscians will bring great danger to Rome. &
    The characters discuss the consequences of banishing Marcius, with Cominius warning that his alliance with the \hlred{Volscians} \hlblue{Romans} will bring great danger to \hlred{Rome} \hlblue{the Volscians}. \texttt{-- Entity Manipulation} \\

    \hline
    \TBstrut
    We introduced a \hlred{novel} \hlblue{new}, simple, and efficient data augmentation method that \hlred{boosts} \hlblue{improves} the performances of existing GANs when training data is limited and diverse. &
    We introduced a novel, simple, and efficient data augmentation method that boosts the performances of existing GANs when training data is \hlred{limited} \hlblue{abundant} and diverse. \texttt{-- Antonym Swap} \\
    
    \hline
    \TBstrut
    Employees of the European Commission are now \hlred{forced} \hlblue{instructed} to \hlred{delete} \hlblue{remove} TikTok from their work devices, and \hlred{delete} \hlblue{get rid of} it from their personal devices too if they have work-related \hlred{apps} \hlblue{applications} installed. &
    Employees of the European Commission are \hlred{now forced} \hlblue{not required} to delete TikTok from their work devices, \hlred{and delete} \hlblue{but should still remove} it from their personal devices \hlred{too} if they have work-related apps installed. \texttt{-- Hallucinated Fact} \\

    \hline
    \TBstrut
    A conversation between a sales agent and a \hlred{potential client} \hlblue{possible customer}. The sales agent provides information on different home insurance \hlred{plans} \hlblue{options} and pricing, as well as available discounts for clients with good credit scores and other factors. &
    A conversation between a sales agent and a potential client. The sales agent provides information on different home insurance plans \hlred{and}\hlblue{, but not on} pricing\hlred{, as well as} \hlblue{or} available discounts for clients with good credit scores and other factors. \texttt{-- Negation Insertion} \\

    \bottomrule
    \end{tabular}
    }
    \caption{Example edit summaries -- \hlred{deletions}, \hlblue{insertions} -- for four domains of \dataset{} (top-to-bottom: Shakespeare Plays, SciTLDR, News, Sales Call). Inconsistent summaries are labeled with an \texttt{Edit Type} which indicates the type of factual inconsistency created with the document (not shown due to length constraint).}
    \label{table:example_edits}
\end{table*}
\subsection{Design Principles}

Based on the analysis of previous benchmarks, we set several design principles that can help create higher quality factual consistency benchmark:

\begin{enumerate}[label=P\arabic*.,ref=P\arabic*]
    \item \textbf{Binary Classification Task:} In the benchmark, a summary should either be labeled as inconsistent if any factual inconsistency is identified with the document or consistent otherwise, to improve label interpretability.
    \item \textbf{Focus on Factual Consistency:} Summaries in the benchmark should be flawless on aspects unrelated to consistency, such as fluency, coherence, and formatting, to avoid confounding effects on the quality of the benchmark.
    \item \textbf{Reproducibility:} Benchmark labels should not depend on annotator identity, and high annotator agreement should confirm the validity of the benchmark, as well as estimate human performance on the benchmark.
    \item \textbf{Benchmark Diversity:} Inconsistency errors in the benchmark should represent a wide range of errors in realistic textual domains, to increase understanding of model strengths and weaknesses, and better establish gaps in performance between models and human annotators at factual reasoning, if there are any.    
\end{enumerate}

\subsection{Creation Procedure}

We now describe the creation procedure we design for the \dataset{} benchmark with an objective to satisfy the design principles stated above, the procedure is visually introduced in Figure~\ref{fig:summedits_protocol}.

At a high level, the procedure consists of three steps: (1) seed summary verification, (2) generation of summary edits, and (3) annotation of edited summaries.

\textbf{Seed Summary Verification.} Benchmark creators select a small collection of documents in a domain of choice, and a \textit{seed summary} is collected for each document, which can either be human-written or model generated. An annotator answers two questions about each \texttt{(document, seed summary)} tuple: (a) ``Are there any flaws with the summary? (fluency, format, etc.)'', (b) ``Is the summary factually consistent with the document?''. If the annotator identifies a flaw in the summary (e.g., an incomplete or disfluent sentence), or any inconsistency, the tuple is filtered out (\textbf{P2}), otherwise, it proceeds to Step 2.

\textbf{Generation of Edited Summaries.} Once a seed summary has been verified, the second step consists in generating multiple \textit{minor edits} to the summary, which might or might not affect the consistency of the summary. This procedure can be carried out manually, or automatically with an LLM. Proposed edits should be atomic and localized, not entirely rewriting a novel summary. Example edits of summaries are shown in Table~\ref{table:example_edits}.

\textbf{Annotation of Edited Summaries.} The annotator who completed the seed verification task (Step 1) is tasked with reviewing each edited summary and assigning it with one of three labels: (a) \textit{consistent} if an edit does not lead to an inconsistency in the summary, (b) \textit{inconsistent} if the edit modifies the seed summary in a way that introduces a factual inconsistency, (c) \textit{borderline} for any other case such as the edit making the summary unclear, or the edit requiring subjective interpretation.

Crucially, we note that a single annotator should complete both Steps 1 and 3, as once they have invested the time in reading the \texttt{(document, summary seed)} tuple, the time required to judge the consistency of edits is greatly reduced. We also recommend including a large number of edits (e.g., 30 edits) to maximize edit diversity (\textbf{P4}), and encouraging annotators to assign the borderline label if they are unsure about any aspect of an edit, in order to maximize reproducibility (\textbf{P3}).

A benchmark can be formed by retaining edited summaries that are labeled as consistent and inconsistent and filtering out borderline cases.

We note that this procedure only requires a small number of documents and seed summaries, as each seed summary is derived into many edited summaries. This flexibility facilitates the creation of factual consistency benchmarks in application domains that lack such resources, such as legal \cite{kornilova2019billsum} or podcast summarization \cite{clifton2020spotify}.

\section{\dataset{} Benchmark}
\label{section:summedits_benchmark}
\subsection{Benchmark Creation}
\begin{table}[]
    \centering
    \begin{tabular}{lccc}
    \toprule
    \textbf{Domain} & \textbf{N} & \textbf{\%Balance} & \textbf{IAA} \\
    \midrule
    News & 819 & 39.2\% & 0.91 \\
    Podcast & 500 & 32.6\% & 0.91 \\
    Billsum & 853 & 42.3\% & 0.90 \\
    Samsum & 664 & 36.4\% & 0.90 \\
    Shakespeare & 814 & 46.4\% & 0.96 \\
    SciTLDR & 466 & 31.1\% & 0.93 \\
    QMSum & 431 & 42.5\% & 0.92 \\
    ECTSum & 668 & 38.0\% & 0.96 \\
    Sales Email & 613 & 29.2\% & 0.87 \\
    Sales Call & 520 & 33.3\% & 0.93 \\
    \midrule
    \textbf{Overall} & 6,348 & 37.10\% & 0.92 \\
    \bottomrule
    \end{tabular}
    \caption{Statistics of the ten domains included in the \dataset{} benchmark, including the number of samples (\textbf{N}), the percentage of consistent summaries (\textbf{\%Balance}), and the inter-annotator agreement (\textbf{IAA}).}
    \label{table:summedits_stats}
\end{table}
\begin{table*}[ht]
\centering
\renewcommand{\arraystretch}{1.2} 
\resizebox{0.98\textwidth}{!}{%
\begin{tabular}{lccccccccccc}
    \toprule
     Model & Podcast & BillSum & SAMSum & News & Sales C & Sales E & Shkspr & SciTLDR & QMSum & ECTSum & Overall ($\downarrow$) \\
    \hline
     DAE & \cellcolor[rgb]{0.89, 0.69, 0.69}54.9 & \cellcolor[rgb]{0.90, 0.70, 0.70}\cellcolor[rgb]{0.90, 0.70, 0.70}55.1 & \cellcolor[rgb]{0.94, 0.81, 0.81}\cellcolor[rgb]{0.94, 0.81, 0.81}59.5 & \cellcolor[rgb]{0.96, 0.87, 0.87}\cellcolor[rgb]{0.96, 0.87, 0.87}61.7 & \cellcolor[rgb]{0.86, 0.59, 0.59} 50.8 & \cellcolor[rgb]{0.90, 0.70, 0.70}55.0 & \cellcolor[rgb]{0.89, 0.68, 0.68}54.5 & \cellcolor[rgb]{0.90, 0.70, 0.70}55.2 & \cellcolor[rgb]{0.87, 0.62, 0.62}\cellcolor[rgb]{0.87, 0.62, 0.62}52.0 & \cellcolor[rgb]{0.93, 0.79, 0.79}58.6 & \cellcolor[rgb]{0.90, 0.71, 0.71}\cellcolor[rgb]{0.90, 0.71, 0.71}55.7 \\
     SummaC & \cellcolor[rgb]{0.93, 0.79, 0.79}58.5 & \cellcolor[rgb]{0.90, 0.71, 0.71}\cellcolor[rgb]{0.90, 0.71, 0.71}55.7 & \cellcolor[rgb]{0.89, 0.69, 0.69}54.7 & \cellcolor[rgb]{0.96, 0.88, 0.88}62.1 & \cellcolor[rgb]{0.93, 0.80, 0.80}59.0 & \cellcolor[rgb]{0.92, 0.77, 0.77}57.7 & \cellcolor[rgb]{0.93, 0.81, 0.81}59.3 & \cellcolor[rgb]{0.94, 0.82, 0.82}\cellcolor[rgb]{0.94, 0.82, 0.82}59.7 & \cellcolor[rgb]{0.91, 0.74, 0.74}\cellcolor[rgb]{0.91, 0.74, 0.74}\cellcolor[rgb]{0.91, 0.74, 0.74}56.6 & \cellcolor[rgb]{0.98, 0.94, 0.94}64.4 & \cellcolor[rgb]{0.93, 0.79, 0.79}58.8 \\
     QAFactEval & \cellcolor[rgb]{0.98, 0.93, 0.93}64.0 & \cellcolor[rgb]{0.89, 0.68, 0.68}\cellcolor[rgb]{0.89, 0.68, 0.68}54.4 & \cellcolor[rgb]{1.00, 0.99, 0.99}66.3 & \cellcolor[rgb]{0.85, 0.90, 0.98}74.6 & \cellcolor[rgb]{0.97, 0.98, 1.00}68.5 & \cellcolor[rgb]{0.98, 0.94, 0.94}64.2 & \cellcolor[rgb]{0.96, 0.88, 0.88}\cellcolor[rgb]{0.96, 0.88, 0.88}61.9 & \cellcolor[rgb]{0.98, 0.99, 1.00}\cellcolor[rgb]{0.98, 0.99, 1.00}67.5 & \cellcolor[rgb]{0.96, 0.89, 0.89}62.4 & \cellcolor[rgb]{0.89, 0.92, 0.99}72.9 & \cellcolor[rgb]{0.99, 0.97, 0.97}\cellcolor[rgb]{0.99, 0.97, 0.97}65.7 \\
    \hline
     Dav001 & \cellcolor[rgb]{0.88, 0.65, 0.65}53.3 & \cellcolor[rgb]{0.85, 0.57, 0.57}50.2 & \cellcolor[rgb]{0.86, 0.59, 0.59}\cellcolor[rgb]{0.86, 0.59, 0.59}51.0 & \cellcolor[rgb]{0.89, 0.68, 0.68}\cellcolor[rgb]{0.89, 0.68, 0.68}54.4 & \cellcolor[rgb]{0.90, 0.70, 0.70}55.3 & \cellcolor[rgb]{0.87, 0.63, 0.63}52.5 & \cellcolor[rgb]{0.85, 0.56, 0.56}\cellcolor[rgb]{0.85, 0.56, 0.56}\cellcolor[rgb]{0.85, 0.56, 0.56}50.0 & \cellcolor[rgb]{0.86, 0.59, 0.59}\cellcolor[rgb]{0.86, 0.59, 0.59}51.0 & \cellcolor[rgb]{0.85, 0.57, 0.57}50.3 & \cellcolor[rgb]{0.86, 0.59, 0.59}50.9 & \cellcolor[rgb]{0.87, 0.61, 0.61}51.9 \\
    Cohere-cmd-XL & \cellcolor[rgb]{0.86, 0.59, 0.59}51.1 & \cellcolor[rgb]{0.87, 0.63, 0.63}52.7 & \cellcolor[rgb]{0.87, 0.62, 0.62}\cellcolor[rgb]{0.87, 0.62, 0.62}52.0 & \cellcolor[rgb]{0.87, 0.63, 0.63}\cellcolor[rgb]{0.87, 0.63, 0.63}52.6 & \cellcolor[rgb]{0.94, 0.83, 0.83}60.3 & \cellcolor[rgb]{0.94, 0.81, 0.81}\cellcolor[rgb]{0.94, 0.81, 0.81}59.5 & \cellcolor[rgb]{0.85, 0.56, 0.56}\cellcolor[rgb]{0.85, 0.56, 0.56}\cellcolor[rgb]{0.85, 0.56, 0.56}50.0 & \cellcolor[rgb]{0.94, 0.84, 0.84}\cellcolor[rgb]{0.94, 0.84, 0.84}60.5 & \cellcolor[rgb]{0.89, 0.67, 0.67}\cellcolor[rgb]{0.89, 0.67, 0.67}\cellcolor[rgb]{0.89, 0.67, 0.67}53.9 & \cellcolor[rgb]{0.94, 0.84, 0.84}\cellcolor[rgb]{0.94, 0.84, 0.84}60.5 & \cellcolor[rgb]{0.90, 0.70, 0.70}\cellcolor[rgb]{0.90, 0.70, 0.70}55.1 \\
     Vicuna-13b & \cellcolor[rgb]{0.88, 0.64, 0.64}52.8 & \cellcolor[rgb]{0.87, 0.63, 0.63}\cellcolor[rgb]{0.87, 0.63, 0.63}52.6 & \cellcolor[rgb]{0.86, 0.59, 0.59}50.8 & \cellcolor[rgb]{0.97, 0.90, 0.90}63.0 & \cellcolor[rgb]{0.92, 0.78, 0.78}58.1 & \cellcolor[rgb]{0.87, 0.61, 0.61}51.8 & \cellcolor[rgb]{0.90, 0.71, 0.71}55.5 & \cellcolor[rgb]{0.94, 0.82, 0.82}\cellcolor[rgb]{0.94, 0.82, 0.82}59.7 & \cellcolor[rgb]{0.89, 0.67, 0.67}54.0 & \cellcolor[rgb]{0.96, 0.89, 0.89}62.5 & \cellcolor[rgb]{0.90, 0.72, 0.72}56.1 \\
     Claude v1.3 & \cellcolor[rgb]{0.94, 0.82, 0.82}\cellcolor[rgb]{0.94, 0.82, 0.82}59.9 & \cellcolor[rgb]{0.87, 0.62, 0.62}52.1 & \cellcolor[rgb]{0.98, 0.93, 0.93}64.1 & \cellcolor[rgb]{0.97, 0.91, 0.91}63.3 & \cellcolor[rgb]{0.96, 0.87, 0.87}\cellcolor[rgb]{0.96, 0.87, 0.87}61.7 & \cellcolor[rgb]{0.91, 0.74, 0.74}\cellcolor[rgb]{0.91, 0.74, 0.74}\cellcolor[rgb]{0.91, 0.74, 0.74}56.6 & \cellcolor[rgb]{0.92, 0.77, 0.77}58.0 & \cellcolor[rgb]{0.92, 0.76, 0.76}57.6 & \cellcolor[rgb]{0.91, 0.75, 0.75}56.9 & \cellcolor[rgb]{0.98, 0.99, 1.00}67.8 & \cellcolor[rgb]{0.94, 0.82, 0.82}59.8 \\
     Dav002 & \cellcolor[rgb]{0.91, 0.73, 0.73}56.4 & \cellcolor[rgb]{0.89, 0.67, 0.67}\cellcolor[rgb]{0.89, 0.67, 0.67}\cellcolor[rgb]{0.89, 0.67, 0.67}53.9 & \cellcolor[rgb]{0.91, 0.75, 0.75}57.1 & \cellcolor[rgb]{0.96, 0.88, 0.88}\cellcolor[rgb]{0.96, 0.88, 0.88}61.9 & \cellcolor[rgb]{0.99, 0.96, 0.96}65.1 & \cellcolor[rgb]{0.93, 0.80, 0.80}59.1 & \cellcolor[rgb]{0.91, 0.74, 0.74}\cellcolor[rgb]{0.91, 0.74, 0.74}\cellcolor[rgb]{0.91, 0.74, 0.74}56.6 & \cellcolor[rgb]{0.98, 0.95, 0.95}64.6 & \cellcolor[rgb]{0.95, 0.84, 0.84}60.6 & \cellcolor[rgb]{1.00, 0.99, 0.99}66.2 & \cellcolor[rgb]{0.94, 0.83, 0.83}60.1 \\
     Bard & \cellcolor[rgb]{0.85, 0.56, 0.56}\cellcolor[rgb]{0.85, 0.56, 0.56}\cellcolor[rgb]{0.85, 0.56, 0.56}50.0 & \cellcolor[rgb]{0.92, 0.78, 0.78}58.3 & \cellcolor[rgb]{0.95, 0.86, 0.86}61.3 & \cellcolor[rgb]{0.89, 0.92, 0.99}72.8 & \cellcolor[rgb]{0.87, 0.91, 0.99}73.8 & \cellcolor[rgb]{0.96, 0.97, 1.00}69.0 & \cellcolor[rgb]{0.93, 0.78, 0.78}58.4 & \cellcolor[rgb]{0.99, 0.99, 0.99}66.1 & \cellcolor[rgb]{0.89, 0.67, 0.67}\cellcolor[rgb]{0.89, 0.67, 0.67}\cellcolor[rgb]{0.89, 0.67, 0.67}53.9 & \cellcolor[rgb]{0.88, 0.92, 0.99}73.1 & \cellcolor[rgb]{0.97, 0.92, 0.92}63.7 \\
     PaLM2-bison & \cellcolor[rgb]{0.99, 0.98, 0.98} 66.0 &\cellcolor[rgb]{0.96, 0.88, 0.88} 62.0 &\cellcolor[rgb]{0.96, 0.97, 1.00} 69.0 &\cellcolor[rgb]{0.97, 0.98, 1.00} 68.4 &\cellcolor[rgb]{0.86, 0.90, 0.98} 74.5 &\cellcolor[rgb]{0.97, 0.98, 1.00} 68.1 &\cellcolor[rgb]{0.95, 0.87, 0.87} 61.6 &\cellcolor[rgb]{0.79, 0.86, 0.98} 78.1 &\cellcolor[rgb]{0.94, 0.96, 0.99} 70.2 &\cellcolor[rgb]{0.90, 0.93, 0.99} 72.3 &\cellcolor[rgb]{0.96, 0.97, 1.00} 69.0  \\
     Dav003 & \cellcolor[rgb]{0.99, 0.97, 0.97}\cellcolor[rgb]{0.99, 0.97, 0.97}65.7 & \cellcolor[rgb]{0.94, 0.82, 0.82}\cellcolor[rgb]{0.94, 0.82, 0.82}59.9 & \cellcolor[rgb]{0.98, 0.99, 1.00}\cellcolor[rgb]{0.98, 0.99, 1.00}67.5 & \cellcolor[rgb]{0.92, 0.94, 0.99}71.2 & \cellcolor[rgb]{0.78, 0.85, 0.97}78.8 & \cellcolor[rgb]{0.95, 0.97, 0.99}69.4 & \cellcolor[rgb]{0.95, 0.96, 0.99}69.6 & \cellcolor[rgb]{0.86, 0.90, 0.98}74.4 & \cellcolor[rgb]{0.90, 0.93, 0.99}72.2 & \cellcolor[rgb]{0.79, 0.86, 0.98}77.9 & \cellcolor[rgb]{0.93, 0.95, 0.99}70.7 \\
     GPT3.5-turbo & \cellcolor[rgb]{0.97, 0.98, 1.00}68.4 & \cellcolor[rgb]{0.97, 0.92, 0.92}63.6 & \cellcolor[rgb]{0.96, 0.97, 0.99}69.1 & \cellcolor[rgb]{0.86, 0.90, 0.98}74.5 & \cellcolor[rgb]{0.76, 0.84, 0.97}79.7 & \cellcolor[rgb]{0.99, 0.97, 0.97}65.5 & \cellcolor[rgb]{0.97, 0.98, 1.00}68.1 & \cellcolor[rgb]{0.84, 0.89, 0.98}75.6 & \cellcolor[rgb]{0.95, 0.97, 0.99}69.2 & \cellcolor[rgb]{0.78, 0.85, 0.97}78.9 & \cellcolor[rgb]{0.92, 0.94, 0.99}71.3 \\
     GPT4 & \cellcolor[rgb]{0.70, 0.79, 0.97}\cellcolor[rgb]{0.70, 0.79, 0.97}83.3 & \cellcolor[rgb]{0.92, 0.94, 0.99}71.1 & \cellcolor[rgb]{0.70, 0.80, 0.97}82.9 & \cellcolor[rgb]{0.70, 0.79, 0.97}\cellcolor[rgb]{0.70, 0.79, 0.97}83.3 & \cellcolor[rgb]{0.65, 0.72, 0.92}87.6 & \cellcolor[rgb]{0.75, 0.83, 0.97}80.1 & \cellcolor[rgb]{0.68, 0.77, 0.95}84.6 & \cellcolor[rgb]{0.71, 0.80, 0.97}\cellcolor[rgb]{0.71, 0.80, 0.97}82.4 & \cellcolor[rgb]{0.75, 0.83, 0.97}80.4 & \cellcolor[rgb]{0.64, 0.71, 0.91}88.0 & \cellcolor[rgb]{0.71, 0.80, 0.97}\cellcolor[rgb]{0.71, 0.80, 0.97}82.4 \\
    \hline
     GPT4 Oracle &     \cellcolor[rgb]{0.61, 0.67, 0.89}90.2 &     \cellcolor[rgb]{0.67, 0.75, 0.94}\cellcolor[rgb]{0.67, 0.75, 0.94}85.5 &    \cellcolor[rgb]{0.66, 0.74, 0.93}\cellcolor[rgb]{0.66, 0.74, 0.93}86.3 &   \cellcolor[rgb]{0.64, 0.70, 0.91}88.3 &         \cellcolor[rgb]{0.60, 0.65, 0.88}91.1 &         \cellcolor[rgb]{0.69, 0.79, 0.96}\cellcolor[rgb]{0.69, 0.79, 0.96}83.5 &         \cellcolor[rgb]{0.54, 0.56, 0.81}96.6 &     \cellcolor[rgb]{0.66, 0.74, 0.93}\cellcolor[rgb]{0.66, 0.74, 0.93}86.3 &    \cellcolor[rgb]{0.62, 0.68, 0.89}89.9 &    \cellcolor[rgb]{0.60, 0.64, 0.87}91.7 &     \cellcolor[rgb]{0.63, 0.69, 0.90}88.9 \\
    
     Human Perf. & \cellcolor[rgb]{0.61, 0.66, 0.88}90.8 & \cellcolor[rgb]{0.65, 0.72, 0.92}87.5 & \cellcolor[rgb]{0.62, 0.68, 0.90}89.4 & \cellcolor[rgb]{0.62, 0.67, 0.89}90.0 & \cellcolor[rgb]{0.60, 0.64, 0.87}91.8 & \cellcolor[rgb]{0.65, 0.72, 0.92}87.4 & \cellcolor[rgb]{0.54, 0.55, 0.81} 96.9 & \cellcolor[rgb]{0.63, 0.69, 0.90}89.3 & \cellcolor[rgb]{0.61, 0.66, 0.88}90.7 & \cellcolor[rgb]{0.55, 0.58, 0.83}95.4 & \cellcolor[rgb]{0.61, 0.66, 0.88}90.9 \\
    \bottomrule
    \end{tabular}
    }
    \caption{Balanced accuracy of models on the \dataset{} benchmark. The top three models are non-LLM specialized models, the middle section are LLMs. We also report a GPT4 oracle performance and an estimate of human performance.}
    \label{table:summedits_results}
\end{table*}

We implemented the \dataset{} protocol on ten realistic summarization domains to explore the reliability of the protocol. For five domains, seed summaries are automatically generated due to the lack or low quality of existing reference summaries. In such cases, we used ChatGPT and domain-specific prompts to generate seed summaries. We note that for all domains, the quality of seed summaries is ultimately manually confirmed in step 1 of the protocol, which consists of ensuring seed summaries are factually consistent and flawless in terms of fluency, formatting, etc.

For all domains, we use GPT3.5-turbo as the LLM to produce edited summaries\footnote{The prompts we use are listed in our open-source release.}. The model chosen to produce summary edits has an important impact on the benchmark. We experimented with integrating multiple LLMs in the edit generation process, but preliminary results indicated that many LLMs were not successful at generating minorly edited summaries and often attempted to write entirely novel summaries, which led us to use ChatGPT as the single model to generate edited summaries. This choice is discussed further in Section~\ref{section:discussion}.

We hired two professional annotators, who were compensated at a rate of \$20/hour to perform steps 1 and 3 of the protocol. Three authors of the paper also participated in the annotation for quality control purposes. Appendix~\ref{appendix:summedits_guidelines} has further detail on annotation protocol, and an overview of the annotation interface, which ensured that each annotator completed Task 1 and 3 sequentially for any sample in the benchmark. We next introduce the ten domains included in the \dataset{} benchmark.

\paragraph{News} To avoid selecting documents and summaries that are in the training corpora of evaluated models, we follow prior work \cite{goyal2022news} and select \texttt{(document, summary)} tuples from recent news articles. We obtained news articles from the Google News top events feed in February 2023, selecting at most one sample per news source to increase coverage diversity \cite{laban2023designing}. Seed summaries are extracted from the article's metadata.

\paragraph{Podcast \cite{clifton2020spotify}} We collected 40 podcast transcripts from the unreleased test set of Spotify's podcast summarization dataset. Due to low reference summary quality, we generated seed summaries automatically.

\paragraph{BillSum \cite{kornilova2019billsum}} We collected 40 US bills and their accompanying summaries as seeds from the training portion of BillSum, a challenging dataset for summarization in the legal domain.

\paragraph{SamSum \cite{gliwa2019samsum}} We collected 40 dialogues and their accompanying summaries from the training portion of SamSum, a common dialogue summarization dataset for messenger-like conversations.

\paragraph{Shakespeare \cite{tiny_shakespeare}} We collected 40 scenes from Shakespeare plays from the Tiny Shakespeare corpus, each roughly 700 words long. We generated seed summaries automatically.

\paragraph{SciTLDR \cite{cachola2020tldr}} We collected 40 research paper abstracts and their corresponding TLDRs from the training portion of SciTLDR, a dataset for scientific paper summarization.

\paragraph{QMSum \cite{zhong2021qmsum}} We collected 40 document and seed summaries from QMSum, a dataset for query-based meeting summarization.

\paragraph{ECTSum \cite{mukherjee2022ectsum}} We collected 40 documents from the ECTSum dataset, a summarization dataset for the financial earnings call transcripts. Due to low reference summary quality, we generated seed summaries automatically.

\paragraph{Sales Call \& Email} We generated fictional sales call transcripts and sales emails -- 40 for each -- and corresponding seed summaries using ChatGPT. This subset of the benchmark evaluates the protocol's validity with textual data entirely machine-generated in targeted domains that lack pre-existing summarization datasets.

\subsection{\dataset{} Statistics}

Table~\ref{table:summedits_stats} provides statistics of the finalized \dataset{} benchmark. Each domain yielded between 400-900 edited summaries, depending on the fraction of seed summaries that pass the first step validation (58\% overall pass rate) and the percentage of edited summaries that are annotated as borderline and filtered out (around 6\%). In the five domains where seed summaries were generated by ChatGPT, 17.8\% of the seed summaries were labeled as factually inconsistent, indicating that modern LLMs like ChatGPT still struggle to remain consistent when summarizing documents.

At least 20\% of each domain's samples were annotated by multiple annotators, allowing us to measure the agreement level when completing the annotation. When considering all three labels (consistent, inconsistent, borderline), Cohen's Kappa in each domain varies between 0.72-0.90, averaging 0.82. When removing samples annotated as borderline by any annotator, the average Cohen's Kappa rises to 0.92, \textbf{empirically validating the importance of labeling and filtering out borderline samples to create a reproducible benchmark.}

In the final benchmark, 37\% of summaries are consistent, and the rest are inconsistent, approaching our objective of a balanced benchmark to facilitate robust evaluation and minimize metric fluctuations \cite{luque2019impact}.

The total annotation cost of \dataset{} is around USD 3,000, representing around 150 hours of annotator work. The average cost of adding a domain to \dataset{} is therefore around USD 300, within reach for NLP practitioners looking to evaluate the model's ability to detect factual errors in their domain of choice. Authors of the FRANK benchmark  \cite{pagnoni2021understanding} -- samples of which are in AggreFact -- estimate that each sample in their benchmark required 30 minutes of annotator time. At similar annotator pay, annotating a benchmark for a new domain similar to the ones in SummEdits would cost an estimated USD 6,000: twenty times more. This cost analysis reveals the dual advantage of our protocol: by focusing the annotation task on atomic edits, costs can be drastically reduced while maintaining high reproducibility.

\subsection{\dataset{} Results}

Table~\ref{table:summedits_results} reports the average performance of specialized models, LLMs with a zero-shot prompt, an oracle version for the LLM in which it has access to additional information and an estimate of human performance computed on the subset of the benchmark which was plurally annotated.

Overall, model performance on the benchmark is low, with a single model -- GPT4 -- getting within 10\% of human performance. Larger or more recent LLMs perform better on the benchmark, illustrated by the performance of models in the OpenAI family, with each model generation leading to an improvement in performance and confirming that the \dataset{} benchmark assesses model ability at factual reasoning.

PaLM2-Bison, Dav003, ChatGPT, and GPT4 are the only four LLMs that outperform the best non-LLM approach QAFactEval, \textbf{providing evidence that most LLMs are not yet capable to reason out-of-the-box about the consistency of facts}.

All three specialized models achieve their highest performance in the news domain, unlike LLM models. The specialized models are likely calibrated to the news domain, which they are most frequently tested on \cite{goyal2020evaluating,laban2022summac,tang2022understanding,fabbri2022qafacteval,}. This finding confirms the importance of creating multi-domain benchmarks to measure model ability in diverse and realistic scenarios.

Some domains such as Shakespeare's plays or the legal BillSum are more challenging to the majority of models, with the latter seeing no model score higher than 71.1\%. Yet, factual reasoning in the legal domain is an important application area of NLP \cite{Chalkidis2020LEGALBERTTM,shen2022multi}.

% Oracle performance
To assess the feasibility of the benchmark, we experiment with an oracle setting of the benchmark, in which the model is provided the seed summary in addition to the input document and the seed summary. The seed summary serves as an information scaffold, enabling the model to directly assess the modifications between the seed and edited summaries when assessing factual consistency. The oracle setting leads to a large boost in performance for the GPT4 model across domains, with the model performing within 2\% of human performance. The GPT4 oracle experiment confirms that high model performance on the benchmark is attainable and that the challenge of \dataset{} lies in aligning the facts of the edited summary with the document, without knowing that it has been edited.

\subsection{Edit Type Analysis}
\label{section:edit_types}
\begin{table}
    \renewcommand{\arraystretch}{1.2} 
    \resizebox{0.48\textwidth}{!}{%
    \begin{tabular}{lcccc}
     & \multicolumn{4}{c}{\textbf{Inconsistent Edit Type}} \\
    \cmidrule{2-5}
     \textbf{Model} & EntMod & Anto & Hallu & Neg \\
    \midrule
     DAE & \cellcolor[rgb]{0.87, 0.62, 0.62} 52.0 & \cellcolor[rgb]{0.88, 0.64, 0.64} 53.0 & \cellcolor[rgb]{0.88, 0.64, 0.64} 52.9 & \cellcolor[rgb]{0.89, 0.67, 0.67} 53.9 \\
     SummaC & \cellcolor[rgb]{0.91, 0.74, 0.74} 56.8 & \cellcolor[rgb]{0.91, 0.74, 0.74} 56.8 & \cellcolor[rgb]{0.90, 0.70, 0.70} 55.3 & \cellcolor[rgb]{0.92, 0.76, 0.76} 57.3 \\
     QAFactEval & \cellcolor[rgb]{0.95, 0.86, 0.86} 61.4 & \cellcolor[rgb]{0.98, 0.96, 0.96} 65.0 & \cellcolor[rgb]{0.98, 0.94, 0.94} 64.3 & \cellcolor[rgb]{0.93, 0.95, 0.99} 70.4 \\ 
     \hline
     Dav001 & \cellcolor[rgb]{0.85, 0.56, 0.56} 50.0 & \cellcolor[rgb]{0.86, 0.59, 0.59} 50.9 & \cellcolor[rgb]{0.86, 0.59, 0.59} 50.8 & \cellcolor[rgb]{0.88, 0.66, 0.66} 53.7 \\
     Cohere-cmd-XL & \cellcolor[rgb]{0.88, 0.66, 0.66} 53.7 & \cellcolor[rgb]{0.90, 0.72, 0.72} 55.8 & \cellcolor[rgb]{0.90, 0.71, 0.71} 55.5 & \cellcolor[rgb]{0.97, 0.93, 0.93} 63.8 \\
     Vicuna-13b & \cellcolor[rgb]{0.90, 0.70, 0.70} 55.2 & \cellcolor[rgb]{0.91, 0.75, 0.75} 57.1 & \cellcolor[rgb]{0.91, 0.73, 0.73} 56.2 & \cellcolor[rgb]{0.95, 0.85, 0.85} 61.0 \\
     Claude v1.3 & \cellcolor[rgb]{0.93, 0.79, 0.79} 58.8 & \cellcolor[rgb]{0.94, 0.83, 0.83} 60.3 & \cellcolor[rgb]{0.95, 0.87, 0.87} 61.5 & \cellcolor[rgb]{1.00, 1.00, 1.00} 66.7 \\
     Dav002 & \cellcolor[rgb]{0.92, 0.78, 0.78} 58.3 & \cellcolor[rgb]{0.95, 0.86, 0.86} 61.4 & \cellcolor[rgb]{0.96, 0.89, 0.89} 62.4 & \cellcolor[rgb]{0.90, 0.93, 0.99} 72.0 \\
     Bard & \cellcolor[rgb]{0.97, 0.91, 0.91} 63.2 & \cellcolor[rgb]{0.99, 0.96, 0.96} 65.3 & \cellcolor[rgb]{0.99, 0.97, 0.97} 65.6 & \cellcolor[rgb]{0.92, 0.94, 0.99} 71.3 \\
     PaLM2-Bison &\cellcolor[rgb]{0.99, 1.00, 1.00} 67.0 &\cellcolor[rgb]{0.94, 0.96, 0.99} 70.0 &\cellcolor[rgb]{0.91, 0.94, 0.99} 71.7 &\cellcolor[rgb]{0.75, 0.83, 0.97} 80.3 \\
     Dav003 & \cellcolor[rgb]{0.95, 0.97, 0.99} 69.2 & \cellcolor[rgb]{0.92, 0.94, 0.99} 71.1 & \cellcolor[rgb]{0.82, 0.88, 0.98} 76.3 & \cellcolor[rgb]{0.70, 0.79, 0.97} 83.3 \\
     GPT3.5-turbo & \cellcolor[rgb]{0.93, 0.95, 0.99} 70.7 & \cellcolor[rgb]{0.93, 0.95, 0.99} 70.6 & \cellcolor[rgb]{0.86, 0.91, 0.98} 74.2 & \cellcolor[rgb]{0.76, 0.84, 0.97} 79.7 \\
     GPT4 & \cellcolor[rgb]{0.72, 0.80, 0.97} 82.2 & \cellcolor[rgb]{0.73, 0.82, 0.97} 81.3 & \cellcolor[rgb]{0.65, 0.73, 0.92} 87.0 & \cellcolor[rgb]{0.59, 0.63, 0.86} 92.7 \\
     \midrule
     \textbf{Average} &\cellcolor[rgb]{0.95, 0.86, 0.86} 61.4 &\cellcolor[rgb]{0.97, 0.90, 0.90} 62.9 &\cellcolor[rgb]{0.98, 0.93, 0.93} 64.1 &\cellcolor[rgb]{0.94, 0.96, 0.99} 69.7 \\
    \bottomrule
    \end{tabular}
    }
    \caption{Balanced accuracy of models on the \dataset{} benchmark, broken down by type of factual error: Entity Modification (\texttt{EntMod}), Antonyms (\texttt{Anto}), Hallucination (\texttt{Hallu}) and Negation (\texttt{Neg}) insertion.}
    \label{table:summedits_types}
\end{table}

To gain more specific insights into the types of edits present in \dataset{}, we annotated each inconsistent sample in the benchmark with tags of edit types that lead to factual inconsistency.

The four types are: (1) \texttt{Entity Modification} in which an entity or phrase in the summary has been changed in a way that alters the meaning, (2) \texttt{Antonym Swap} is when a word or phrase is replaced by a word of opposite meaning (e.g., increasing vs. decreasing), (3) \texttt{hallucinated fact insertion} is a novel fact is introduced in the summary which is not supported by the document, and (4) \texttt{negation insertion} is the use of any negator word (e.g., not, neither) which modifies summary meaning. Figure~\ref{table:example_edits} provides an example of each edit type in \dataset{}.

To annotate the entire benchmark, one author of the paper first manually annotated 200 samples of the dataset, which was used to evaluate several GPT4-based Zero-Shot and Few-Shot approaches. The best approach was then used to annotate each edited summary with edit types.

The best-performing prompt provides the definition of each edit type and a canonical example of each, and it achieved a performance of 0.85 F-1 and 0.92 recall, which was deemed sufficient for analysis purposes.\footnote{We provide the prompt with the code release.}

Overall in \dataset{}, 78\% of inconsistent summaries contain an entity modification, 48\% an antonym swap, 22\% a hallucinated fact insertion, and 18\% a negator insertion. We note that the distribution of edit types is highly influenced by the LLM used to produce the edits, which is ChatGPT in our case. Table~\ref{table:summedits_types} presents model performance across each of the edit types.

All models detect inconsistencies due to negator insertions the best, a sign that such errors are more discernable to models. Fact hallucinations are relatively harder to detect for non-LLM models but gradually become more evident to more performant LLMs. Finally, the entity modification and antonym error types generally see the lowest rate of detection by models across the board, perhaps due to such edits modifying an existing consistent fact in a more nuanced way.

\subsection{Number of Edits Effect}
\begin{table}
    \centering
\renewcommand{\arraystretch}{1.2} 
\resizebox{0.45\textwidth}{!}{%
\begin{tabular}{lcccc}
 & \multicolumn{4}{c}{\textbf{\#Distinct Edit Types}} \\
\cmidrule{2-5}
 \textbf{Model} & 1 & 2 & 3 & 4 \\
\midrule
 DAE & \cellcolor[rgb]{0.85, 0.57, 0.57} 50.2 & \cellcolor[rgb]{0.88, 0.66, 0.66} 53.5 & \cellcolor[rgb]{0.90, 0.71, 0.71} 55.4 & \cellcolor[rgb]{0.98, 0.95, 0.95} 64.9 \\
 SummaC & \cellcolor[rgb]{0.92, 0.78, 0.78} 58.2 & \cellcolor[rgb]{0.91, 0.73, 0.73} 56.3 & \cellcolor[rgb]{0.92, 0.76, 0.76} 57.6 & \cellcolor[rgb]{0.99, 0.99, 1.00} 67.3 \\
 QAFactEval & \cellcolor[rgb]{0.93, 0.81, 0.81} 59.4 & \cellcolor[rgb]{0.97, 0.92, 0.92} 63.7 & \cellcolor[rgb]{0.90, 0.93, 0.99} 72.3 & \cellcolor[rgb]{0.82, 0.88, 0.98} 76.5 \\
\hline
 Dav001 & \cellcolor[rgb]{0.85, 0.56, 0.56} 50.0 & \cellcolor[rgb]{0.85, 0.58, 0.58} 50.5 & \cellcolor[rgb]{0.89, 0.67, 0.67} 53.9 & \cellcolor[rgb]{0.97, 0.91, 0.91} 63.1 \\
 Vicuna-13b & \cellcolor[rgb]{0.88, 0.64, 0.64} 52.8 & \cellcolor[rgb]{0.91, 0.75, 0.75} 57.0 & \cellcolor[rgb]{0.94, 0.83, 0.83} 60.2 & \cellcolor[rgb]{0.93, 0.79, 0.79} 58.5 \\
 Cohere-cmd-XL & \cellcolor[rgb]{0.85, 0.56, 0.56} 50.0 & \cellcolor[rgb]{0.90, 0.72, 0.72} 55.9 & \cellcolor[rgb]{0.97, 0.92, 0.92} 63.7 & \cellcolor[rgb]{0.94, 0.96, 0.99} 70.0 \\
 Claude v1.3 & \cellcolor[rgb]{0.92, 0.76, 0.76} 57.5 & \cellcolor[rgb]{0.95, 0.84, 0.84} 60.6 & \cellcolor[rgb]{0.99, 0.97, 0.97} 65.4 & \cellcolor[rgb]{0.98, 0.94, 0.94} 64.3 \\
 Dav002 & \cellcolor[rgb]{0.91, 0.73, 0.73} 56.3 & \cellcolor[rgb]{0.95, 0.86, 0.86} 61.2 & \cellcolor[rgb]{0.95, 0.97, 0.99} 69.4 & \cellcolor[rgb]{0.72, 0.81, 0.97} 81.7 \\
 Bard & \cellcolor[rgb]{0.95, 0.85, 0.85} 61.0 & \cellcolor[rgb]{0.98, 0.95, 0.95} 64.9 & \cellcolor[rgb]{0.90, 0.93, 0.99} 72.4 & \cellcolor[rgb]{0.88, 0.92, 0.99} 73.4 \\
 PaLM2-Bison &\cellcolor[rgb]{0.99, 0.99, 0.99} 66.1 &\cellcolor[rgb]{0.95, 0.96, 0.99} 69.5 &\cellcolor[rgb]{0.76, 0.84, 0.97} 79.6 &\cellcolor[rgb]{0.95, 0.97, 0.99} 69.4 \\
 ChatGPT & \cellcolor[rgb]{0.97, 0.98, 1.00} 68.5 & \cellcolor[rgb]{0.91, 0.94, 0.99} 71.4 & \cellcolor[rgb]{0.72, 0.81, 0.97} 82.0 & \cellcolor[rgb]{0.66, 0.73, 0.93} 86.6 \\
 Dav003 & \cellcolor[rgb]{0.99, 0.96, 0.96} 65.3 & \cellcolor[rgb]{0.90, 0.93, 0.99} 72.0 & \cellcolor[rgb]{0.67, 0.75, 0.94} 85.8 & \cellcolor[rgb]{0.63, 0.69, 0.90} 88.8 \\
 GPT4 & \cellcolor[rgb]{0.74, 0.82, 0.97} 81.0 & \cellcolor[rgb]{0.70, 0.79, 0.97} 83.0 & \cellcolor[rgb]{0.59, 0.64, 0.87} 92.0 & \cellcolor[rgb]{0.57, 0.60, 0.84} 94.3 \\
 \midrule
  \textbf{Average} & \cellcolor[rgb]{0.93, 0.81, 0.81} 59.2 & \cellcolor[rgb]{0.96, 0.89, 0.89} 62.5 & \cellcolor[rgb]{0.95, 0.97, 0.99} 69.2 & \cellcolor[rgb]{0.86, 0.91, 0.98} 74.1 \\
\bottomrule
\end{tabular}
}
\caption{Relationship between the number of edits types in the summary and balanced accuracy of models on \dataset{}. Models generally perform better as the number of introduced edits in a summary increases.}
\label{tab:summedits_num_edits}
\end{table}

It is common for the LLM to introduce multiple edits in each of its candidate summaries, as can be seen in the examples in Table~\ref{table:example_edits}, in which each edited summary contains multiple inserted and deleted words. We group inconsistent summaries by the number of distinct edit types they contain (1 to 4) and compute model performance on each group, with results summarized in Table~\ref{tab:summedits_num_edits}.

As the number of edit types in a summary increases, most models see sizable performance improvements, with average performance increasing from 59.2 to 74.1 between summaries with 1 or 4 edit types represented.

This analysis confirms the perspective the task in the \dataset{} benchmark corresponds to a \textit{detection} task: as the number of introduced errors increases, model performance increases as there is generally more evidence of inconsistencies for the models to detect. This also points in the direction of a more challenging explanation analysis, in which one could annotate whether a model can detect all inconsistencies in a summary.

In turn, future work looking to create more challenging versions of benchmarks using the SummEdits protocol can focus on editing summaries with a single edit type, as such inconsistent summaries are more challenging to detect.

\section{Limitations and Discussion}
\label{section:discussion}

\paragraph{Why not fix existing benchmarks?} In Section~\ref{table:existing_benchmarks}, analysis reveals limitations with existing benchmarks that in theory can be fixed to yield improved versions of known benchmarks. The analysis we performed however only helps us invalidate a subset of samples in an opportunistic way, by looking at samples where benchmark labels and GPT4 disagree. However, this methodology cannot help us efficiently correct or confirm all samples, and improving existing benchmarks would require re-annotating a large portion of the benchmarks, and we do not have a guarantee that new annotations would improve on previous ones. By designing a new protocol for sample annotation that relies on clear, atomic edits, we simplify the annotation process, improving reproducibility. 

\paragraph{Effect of LLM in benchmark creation.} Step 2 of the protocol described in Section~\ref{section:protocol} relies on an LLM to generate many edits of the seed summary, which are subsequently manually annotated and included in the benchmark. The choice of LLM likely has an effect on the benchmark which could favor a subset of LLMs most similar to the one used for benchmark creation. Initial attempts to use a pool of LLMs to produce edits were unsuccessful as we found that only ChatGPT and GPT4 were currently capable of following editing instructions that do not fully rewrite summaries. Future iterations on similar benchmarks should consider including diverse pools of LLMs in benchmark creation processes to avoid model-specific bias.

\paragraph{Evalutating Summarizers.} Previous benchmarks were in part collected to evaluate which summarization models are least likely to generate factual inconsistencies \cite{falke2019ranking}. Since the summaries in \dataset{} are synthetic modifications of summaries, the benchmark cannot directly provide insights on summarizers and their ability to remain consistent. Future work can explore using methods such as Near-Negative Distinction (NND) \cite{Laban2022NearNegativeDG} to adapt \dataset{} into a set of tests to evaluate summarizer performance, and model ability to avoid generating inconsistent samples in the first place.

\paragraph{Build Your Own Benchmark.} By implementing the protocol in ten diverse domains for an average cost of around USD300 per domain, we've demonstrated that the protocol can be adapted to widely different textual domains -- from US legal bills to Shakespeare plays -- and produce domain-specific benchmarks. Although we hope that the domains we've selected span a range of practical use cases, we hope that others will adopt and adapt the protocol to new domains, languages, and NLP tasks.

\section{Conclusion}
In this work, we explore the capabilities of LLMs to act as factual reasoners through the lens of factual evaluation in text summarization. We show that on a surface level, LLMs perform on par with specialized non-LLM evaluators, but the performance substantially degrades in more advanced evaluation settings. As part of this analysis, we also uncover and discuss shortcomings of existing benchmarks for factual evaluation. Using those insights we develop a new protocol for creating inconsistency detection benchmarks, which we implement in a 10-domain benchmark called \dataset{}. The \dataset{} benchmark is highly reproducible and more cost-effective per sample than previous benchmarks. Our experiments show that the benchmark is challenging for most current LLMs, with the best-performing model, GPT-4, still 8\% below estimated human performance.
We believe that \dataset{} can serve as a valuable tool for evaluating LLMs' abilities to reason about facts, detect factual errors and promote more reliable NLG systems. We encourage LLM developers to report their performance on the benchmark, and practitioners to adapt the protocol to generate in-domain benchmarks for model evaluation.

\bibliography{custom}
\bibliographystyle{acl_natbib}

\appendix

% \section{Prompt Templates}
% \label{appendix:prompt_templates}
% \input{tables/prompt-examples}

\section{Model Access Detail}
\label{appendix:model_access}

In Section~\ref{section:factcc_analysis}, we experiment with a wide range of models. For each model, we specify its model card, and how it was accessed.

\paragraph{Non-LLM models.} The three specialized models -- SummaC\footnote{\url{https://github.com/tingofurro/summac}}, DAE\footnote{\url{https://github.com/tagoyal/factuality-datasets}}, and QAFactEval\footnote{\url{https://github.com/salesforce/QAFactEval}} -- were implemented through their online public repositories, and run locally on a multi-GPU machine (with 2 V-100 GPUs).

\paragraph{Open-source Models.} We experimented with five open-source LLM models: LLama-13b \cite{touvron2023llama}, Alpaca-13b \cite{taori2023alpaca}, Dolly-V2-12b (\texttt{databricks/dolly-v2-12b}), Vicuna-13b \cite{chiang2023vicuna}, and MosaicML's MPT-7b-chat \cite{MosaicML2023Introducing}. All models were accessed through the public, online demonstration of LMSys.org\footnote{\url{https://chat.lmsys.org/}}. Model responses were collected between April 15th, 2023, and May 15th, 2023.

\paragraph{Google Models.} We experiment with two Google models, the Bard \cite{thoppilan2022lamda} which we accessed through a web-based interface\footnote{\url{https://bard.google.com/}} which does not specify an exact model card, but model responses were collected between April 15th, 2023 and May 15th, 2023. Second, the PaLM-v2-bison model \cite{narang2022pathways} (model card \texttt{text-bison@001}), which was accessed through the Google Cloud VertexAI API.

\paragraph{Anthropic Model.} We collected outputs of the Claude V1.3 model (model card: \texttt{claude-v1.3}), the latest and largest Anthropic model at the time of publication, using the official API hosted by Anthropic\footnote{\url{https://github.com/anthropics/anthropic-sdk-python}}. 

\paragraph{Cohere Model.} We collected outputs of Cohere's \texttt{command-xlarge} model, the latest and largest Cohere model at the time of publication, using the official API hosted by Cohere\footnote{\url{{https://docs.cohere.com/docs/the-cohere-platform}}}.

\paragraph{OpenAI Models.} We collected outputs for eight OpenAI models. Six models are from the GPT-3 family: Ada001 (\texttt{text-ada-001}), Bab001 (\texttt{text-babbage-001}), Cur001 (\texttt{text-curie-001}), Dav001 (\texttt{text-davinci-001}), Dav002 (\texttt{text-davinci-002}), and Dav003 (\texttt{text-davinci-003}). We also include GT3.5-turbo (\texttt{gpt-3.5-turbo}) and GPT-4 (\texttt{gpt-4}). All models were accessed through OpenAI's official API\footnote{\url{https://github.com/openai/opeai-python}}.

\section{Explanation Annotation Guidelines}
\label{appendix:reasoning_guidelines}

We hired two professional annotators to complete the annotation of model-generated explanations on the FactCC and AggreFact domains. The annotators were compensated at \$20/hour. They received onboarding documentation that introduced them to the task, and provided the following definition for each type of explanation:
\begin{itemize}
    \item \textbf{No Explanation}: If the model did not provide any explanation. (For
example just saying: ``The summary is inconsistent''),
    \item \textbf{Entirely Correct}: if the explanation correctly identifies and explains one or more factual inconsistencies in the summary,
    \item \textbf{Partially Correct}: if the explanation provided contains several
elements and at least one of them correctly identifies and explains a factual inconsistency in the summary,
    \item \textbf{Unrelated}: if the explanation given does not directly relate to a factual inconsistency between the summary and the document,
    \item \textbf{Incorrect}: if the explanation given does not correctly identify a
factual inconsistency in the summary, for example, making a logical error.
\end{itemize}

An example for each type of explanation was provided during onboarding, similar to the ones given in Table~\ref{table:example_explanations}. In order to obtain impartial results that do not benefit or disadvantage any model, for cases where multiple explanations were annotated for the same \texttt{(document, summary)} sample, the explanations' order was shuffled, and annotators were not aware of the model origin of any explanation.

Annotation was performed in batches, and the first two batches of annotation of each annotator were reviewed by the authors of the paper. Incorrect annotations were discussed, allowing annotators to better understand edge cases of the task, and modify their annotation in the first batches. The annotators were added to a Slack channel with one of the authors and regularly discussed edge cases to maintain a common understanding of the task. For example, both annotators raised the question of how to deal with cut-off explanations, in which the last sentence is incomplete (due to the max-length of generation). Annotators were both instructed to disregard any incomplete sentence and only consider full sentences in their assessment.

\section{\dataset{} Annotation Guidelines}
\label{appendix:summedits_guidelines}

\begin{figure*}
    \centering
    \fbox{
    \includegraphics[width=0.9\textwidth]{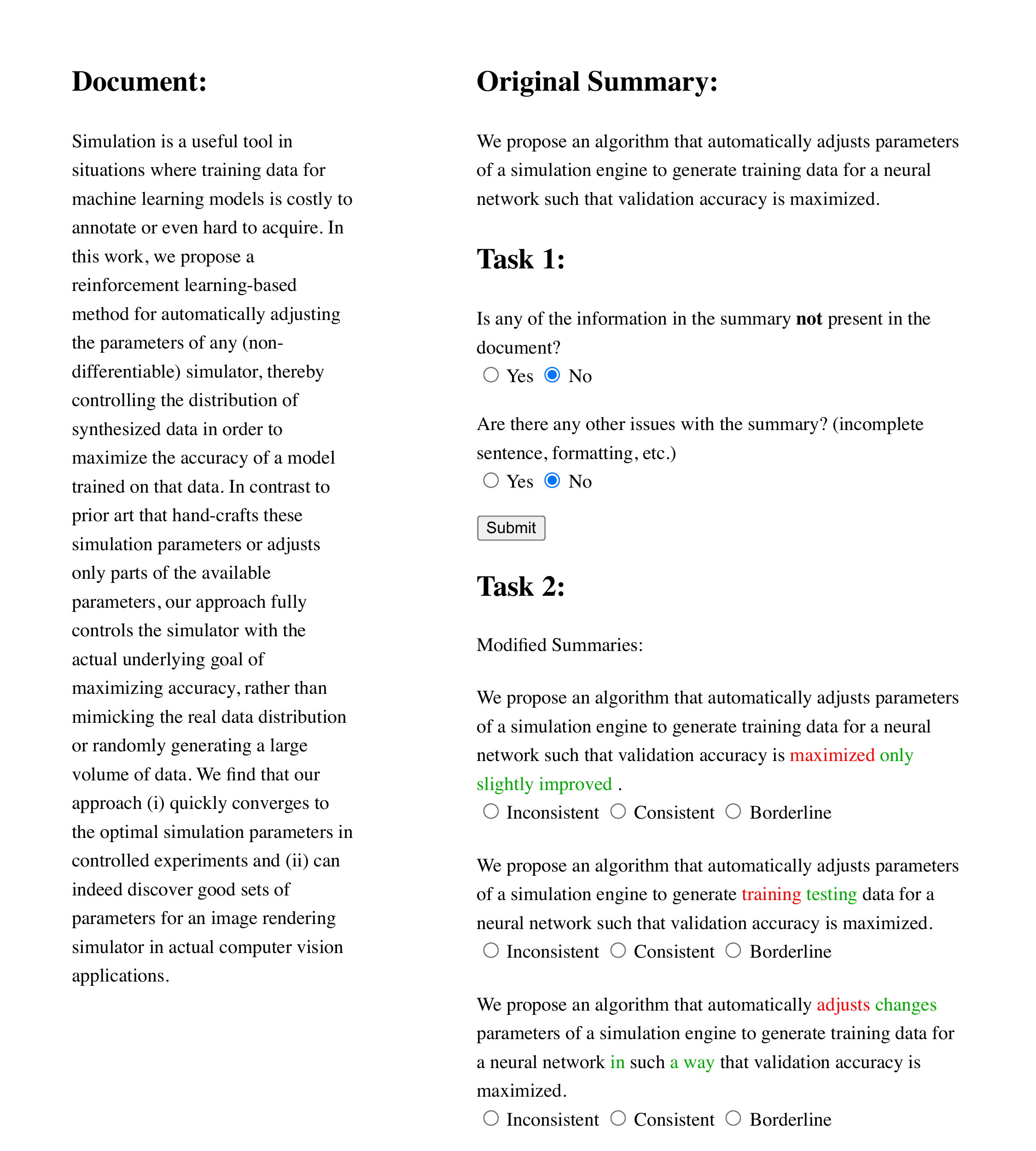}
    }
    \caption{Two-column annotation interface used to annotate samples in the \dataset{} benchmark. Participants could read the document on the left-hand column. Once they completed Task 1 in the right-hand column, the second annotation task became visible.}
    \label{fig:summedits_interface}
\end{figure*}

We hired two professional annotators to complete the annotation of Steps 1 and 3 of the \dataset{} protocol (see Section~\ref{section:protocol}). The annotators were compensated at \$20/hour. They received onboarding documentation that introduced them to the task and used the interface shown in Figure~\ref{fig:summedits_interface}.

Annotators were first assigned 10 warm-up seed summaries, each with roughly 30 edited summaries, which had been pre-annotated by the authors of the paper. The authors reviewed the completed warm-up exercises, and a strong agreement level on the warm-up task with both annotators was observed. We discussed disagreement cases with the annotators and added both annotators to a Slack channel with one of the authors of the paper to allow them to discuss any edge case or domain-specific question. For example, since the QMSumm domain is the more specific query-focused summarization, the annotators were given updated instructions on Slack on how to deal with the ``query'' component when evaluating summaries. Namely, during Step 1 of the protocol, participants were asked to additionally judge whether the summary accurately responded to the query, and otherwise mark summaries as inadequate.

\end{document}